\documentclass[twoside,palatino,psgreek]{article}
\usepackage{coli}
\usepackage{fullname}
\usepackage{url}
\usepackage{haldefs}
\usepackage{psfig}

\title{Induction of Word and Phrase Alignments for Automatic Document Summarization}

\author{ Hal Daum\'e III and Daniel Marcu%
         \thanks{4676 Admiralty Way, Suite 1001, Marina del Rey, CA 90292.%
         ~~Email: {\tt \{hdaume,marcu\}@isi.edu}%
         \protect\linebreak\protect\linebreak%
         Submission received: 12 January 2005; Revised submission%
         received: 3 May 2005; Accepted for publication: 27 May 2005}\\
         \affil{Information Sciences Institute} \\ \affil{University of Southern California}}

\runningtitle{Alignments for Document Summarization}
\runningauthor{Daum\'e III \& Marcu}

\newcommand{\jump}{\textrm{jump}}
\newcommand{\rewrite}{\textrm{rewrite}}

\renewcommand{\O}[1]{\mathcal{O}\left(#1\right)}

\newcommand{\docabs}{$\langle$document, abstract$\rangle$}
\newcommand{\extabs}{$\langle$extract, abstract$\rangle$}

\begin{document}

\setlength{\baselineskip}{1\baselineskip}      

\maketitle

\begin{abstract}
\setlength{\baselineskip}{1\baselineskip}      
Current research in automatic single document summarization is
dominated by two effective, yet na\"ive approaches: summarization by
sentence extraction, and headline generation via bag-of-words models.
While successful in some tasks, neither of these models is able to
adequately capture the large set of linguistic devices utilized by
humans when they produce summaries.  One possible explanation for the
widespread use of these models is that good techniques have been
developed to extract appropriate training data for them from existing
document/abstract and document/headline corpora.  We believe that
future progress in automatic summarization will be driven both by the
development of more sophisticated, linguistically informed models, as
well as a more effective leveraging of document/abstract corpora.  In
order to open the doors to simultaneously achieving both of these
goals, we have developed techniques for automatically producing
word-to-word and phrase-to-phrase \emph{alignments} between documents
and their human-written abstracts.  These alignments make explicit the
correspondences that exist in such document/abstract pairs, and create
a potentially rich data source from which complex summarization
algorithms may learn.  This paper describes experiments we have
carried out to analyze the ability of \emph{humans} to perform such
alignments, and based on these analyses, we describe experiments for
creating them automatically.  Our model for the alignment task is
based on an extension of the standard hidden Markov model, and learns
to create alignments in a completely unsupervised fashion.  We
describe our model in detail and present experimental results that
show that our model is able to learn to reliably identify word- and
phrase-level alignments in a corpus of \docabs\ pairs.
\end{abstract}

\section{Introduction and Motivation} \label{intro}

\subsection{Motivation}

We believe that future success in automatic document summarization
will be made possible by the combination of complex, linguistically
motivated models and effective leveraging of data.  Current research
in summarization makes a choice between these two: one either develops
sophisticated, domain-specific models that are subsequently hand-tuned
without the aid of data, or one develops na\"ive general models that
can be trained on large amounts of data (in the form of corpora of
document/extract or document/headline pairs).  One reason for this is
that currently available technologies are only able to extract very
coarse and superficial information that is inadequate for training
complex models.  In this paper, we propose a method to overcome this
problem, by automatically generating word-to-word and phrase-to-phrase
\emph{alignments} between documents and their human-written
abstracts.\footnote{\setlength{\baselineskip}{1\baselineskip}We will use the words ``abstract'' and ``summary''
interchangeably.  When we wish to emphasize that a particular summary
is \emph{extractive}, we will refer to it as an ``extract.''}

\begin{figure}[!bt]
\mbox{\psfig{figure=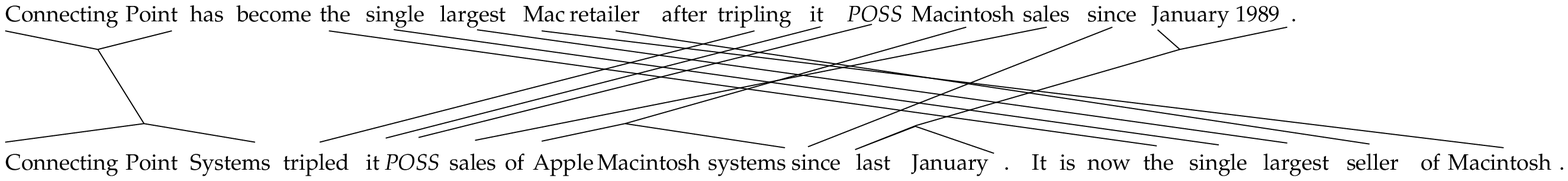,width=33pc}}
\caption{\setlength{\baselineskip}{1\baselineskip}Example alignment of a single abstract sentence with two
  document sentences.}
\label{alignment-ex}
\end{figure}

To facilitate discussion and to motivate the problem, we show in
Figure~\ref{alignment-ex} a relatively simple alignment between a
document fragment and its corresponding abstract fragment from our
corpus.  In this example, a single abstract sentence (shown along the
top of the figure) corresponds to exactly two document sentences
(shown along the bottom of the figure).  If we are able to
automatically generate such alignments, one can envision the
development of models of summarization that take into account effects
of word choice, phrasal and sentence reordering, and content
selection.  Such models could be simultaneously linguistically
motivated and data-driven.  Furthermore, such alignments are
potentially useful for current-day summarization techniques, including
sentence extraction, headline generation and document compression.

A close examination of the alignment shown in
Figure~\ref{alignment-ex} leads us to three observations about the
nature of the relationship between a document and its abstract, and
hence about the alignment itself:

\begin{itemize}
\item Alignments can occur at the granularity of words and of phrases.
\item The ordering of phrases in an abstract can be different from the
  ordering of phrases in the document.
\item Some abstract words do not have direct correspondents in the
  document, and many document words are never used in an abstract.
\end{itemize}

In order to develop an alignment model that could potentially recreate
such an alignment, we need our model to be able to operate both at the
word level and at the phrase level, we need it to be able to allow
arbitrary reorderings, and we need it to be able to account for words
on \emph{both} the document \emph{and} abstract side that have no
direct correspondence.  In this paper, we develop an alignment model
that is capable of learning all these aspects of the alignment problem
in a completely unsupervised fashion.

\subsection{Shortcomings of current summarization models}

Current state of the art automatic single-document summarization
systems employ one of three techniques: sentence extraction,
bag-of-words headline generation, or document compression.  Sentence
extraction systems take full sentences from a document and concatenate
them to form a summary.  Research in sentence extraction can be traced
back to work in the mid fifties and late sixties by
Luhn~\shortcite{luhn56} and Edmundson~\shortcite{edmundson69}.  Recent
techniques are startlingly not terribly divergent from these original
methods; see \cite{mani-maybury-book99,marcu-book00,mani-book01} for a
comprehensive overview.  Headline generation systems, on the other
hand, typically extract individual words from a document to produce a
very short headline-style summary; see
\cite{bankoetal00,berger-mittal00,schwartz-headline} for
representative examples.  Between these two extremes, there has been a
relatively modest amount of work in sentence simplification
\cite{chandrasekaretal96,mahesh97,carrolletal98,grefenstette98,jing00,knight-marcu02}
and document compression
\cite{daume02noisychannel,daume04treeposition,zajic04topiary} in which
words, phrases and sentences are selected in an extraction process.

While such approaches have enjoyed some success, they all suffer from
modeling shortcomings.  Sentence extraction systems and document
compression models make unrealistic assumptions about the
summarization task (namely, that extraction is sufficient, and that
sentences are the appropriate level of granularity).  Headline
generation systems employ very weak models that make an incorrect
bag-of-words assumption.  This assumption allows such systems to learn
limited transformations to produce headlines from arbitrary documents,
but such transformations are not nearly complex enough to adequately
model anything beyond indicative summaries at a length of around ten
words.  Bag of words models can learn what the most important words to
keep in a headline are, but say nothing about how to structure them in
a well-formed, grammatical headline.

In our own work on document compression models
\cite{daume02noisychannel,daume04treeposition}, both of which extend
the sentence compression model of Knight and
Marcu~\shortcite{knight-marcu02}, we assume that sentences and
documents can be summarized exclusively through deletion of contiguous
text segments.  In Knight and Marcu's data, we found that from a
corpus of $39,060$ abstract sentences, only $1067$ were created from
corresponding document sentences via deletion of contiguous segments.
In other words, only $2.7\%$ of the sentences in real \docabs\ pairs
can be explained by the model proposed by Knight and Marcu
\shortcite{knight-marcu02}.  Such document compression models do not
explain the rich set of linguistic devices employed, for example, in
Figure~\ref{alignment-ex}.

\subsection{Prior work on alignments}

In the sentence extraction community, there exist a wide variety of
techniques for (essentially) creating alignments between document
\emph{sentences} and abstract \emph{sentences}
\cite{kupiec95trainable,teufel97sentence,marcu-sigir99}; see also
\cite{barzilay03alignment,quirk04alignment} for work describing
alignments for the monolingual paraphrasing task.  These techniques
typically take into account information such as lexical overlap,
synonymy, ordering, length, discourse structure, and so forth.  The
sentence alignment problem is a comparatively simple problem to solve
and current approaches work quite well.  Unfortunately, these
alignments are the least useful, as they can only be used to train
sentence extraction systems.

In the context of headline generation, simple statistical models are
used for aligning documents and headlines
\cite{bankoetal00,berger-mittal00,schwartz-headline}, based on IBM
Model 1~\cite{brownetal93}.  These models treat documents and
headlines as simple bags of words and learn probabilistic word-based
mappings between the words in the documents and the words in the
headlines.  Such mappings can be considered word-to-word alignments,
but as our results show (see Section~\ref{sec:results}), these models
are too weak for capturing the sophisticated operations that are
employed by humans in summarizing texts.

To date, there has been very little work on the word alignment task in
the context of summarization.  The most relevant work is that of Jing
\shortcite{jing:cl}, in which a hidden Markov alignment model is
applied to the task of identifying word and phrase-level
correspondences between documents and abstracts.  Unfortunately, this
model is only able to align words that are identical up to their
stems, and thus suffers from a problem of recall.  This also makes it
ill-suited to the task of learning how to perform abstraction, in
which one would desire to know how words get changed.  For example,
Jing's model cannot identify any of the following alignments from
Figure~\ref{alignment-ex}: (Connecting Point $\leftrightarrow$
Connecting Point Systems), (Mac $\leftrightarrow$ Macintosh),
(retailer $\leftrightarrow$ seller), (Macintosh $\leftrightarrow$
Apple Macintosh systems) and (January 1989 $\leftrightarrow$ last
January).

Word alignment (and, to a lesser degree, phrase alignment) has been an
active topic of research in the machine translation community.  Based
on these efforts, one might be initially tempted to use
readily-available alignment models developed in the context of machine
translation, such as GIZA++ \cite{och03} to obtain word-level
alignments in \docabs\ corpora.  However, as we will show
(Section~\ref{sec:results}), the alignments produced by such a system
are inadequate for the \docabs\ alignment task.

\subsection{Paper structure}

In this paper, we describe a novel, general model for automatically
inducing word- and phrase-level alignments between documents and their
human-written abstracts.  Beginning in Section~\ref{sec:human}, we
will describe the results of \emph{human} annotation of such
alignments.  Based on this annotation, we will investigate the
empirical linguistic properties of such alignments, including lexical
transformations and movement.  In Section~\ref{sec:model}, we will
introduce the statistical model we use for deriving such alignments
automatically.  The inference techniques are based on those of
semi-Markov models, extensions of hidden Markov models that allow for
multiple simultaneous observations.

After our discussion of the model structure and algorithms, we discuss
the various parameterizations we employ in
Section~\ref{sec:parameterization}.  In particular, we discuss three
distinct models of movement, two of which are well-known in the
machine translation alignment literature, and a third one that
exploits syntax in a novel, ``light'' manner.  We also discuss several
models of lexical rewrite, based on identities, stems, WordNet
synonymy and automatically induced lexical replacements.  In
Section~\ref{sec:results}, we present experimental results that
confirm that our model is able to learn the hidden structure in our
corpus of \docabs\ pairs.  We compare our model against well-known
alignment models designed for machine translation, as well as a state
of the art alignment model specifically designed for summarization
\cite{jing:cl}.  Additionally, we discuss errors that the model
currently makes, supported by some relevant examples and statistics.
We conclude with some directions for future research
(Section~\ref{sec:conclusion}).

\section{Human-produced Alignments} \label{sec:human}

\begin{table}[tb]
\tcaption{\setlength{\baselineskip}{1\baselineskip}Ziff-Davis corpus statistics}
\label{corpus-statistics}
\begin{tabular}{|l||r|r||r|r|}
\hline
& \multicolumn{2}{c||}{\bf Sub-corpus} & \multicolumn{2}{c|}{\bf Annotated} \\
&           {\bf Abstracts} & {\bf Documents}  
&           {\bf Abstracts} & {\bf Documents}  \\
\hline
Documents           &\multicolumn{2}{c||}{$2033$}
                    &\multicolumn{2}{c|}{$45$}\\
Sentences           &   $13k$ &           $82k$ & $244$ & $2k$ \\
Words               &  $261k$ &        $2.5m$ & $6.4k$ & $49k$ \\
\hline
Unique words               &   $14k$ &           $42k$ & $1.9k$ &
$5.9k$ \\
                    &\multicolumn{2}{c||}{$45k$}  &\multicolumn{2}{c|}{$6k$}    \\
\hline
Sentences/Doc       &     $6.28$ &            $40.83$ & $5.42$ & $45.3$\\
Words/Doc           &   $128.52$ &           $1229.71$ & $142.33$ & $1986.16$ \\
\hline
Words/Sent          &    $20.47$ &            $28.36$ & $26.25$ & $24.20$ \\
\hline
\end{tabular}
\end{table}

In order to decide how to design an alignment model and to judge the
quality of the alignments produced by a system, we first need to
create a set of ``gold standard'' alignments.  To this end, we asked
two human annotators to manually construct such alignments between
documents and their abstracts.  These \docabs\ pairs were drawn from
the Ziff-Davis collection \cite{marcu-sigir99}.  Of the roughly $7000$
documents in that corpus, we randomly selected $45$ pairs for
annotation.  We added to this set of $45$ pairs the $2000$ shorter
documents from this collection, and all the work described in the
remainder of this paper focuses on this subset of $2033$ \docabs\
pairs\footnote{\setlength{\baselineskip}{1\baselineskip}The reason there are $2033$ pairs, not $2045$, is that
12 of the original 45 pairs were among the $2000$ shortest, so the
$2033$ pairs are obtained by taking the $2000$ shortest and adding to
them the $33$ pairs that were annotated and not already among the
$2000$ shortest.}.  Statistics for this sub-corpus and for the pairs
selected for annotation are shown in Table~\ref{corpus-statistics}.
As can be simply computed from this table, the compression rate in
this corpus is about $12\%$.  The first five human-produced alignments
were completed separately and then discussed; the last 40 were done
independently.

\subsection{Annotation guidelines}

Annotators were asked to perform word-to-word and phrase-to-phrase
alignments between abstracts and documents and to classify each
alignment as either possible ($P$) or sure ($S$), where $S \subseteq P$,
following the methodology used in the machine translation community
\cite{och03}.  The direction of containment ($S \subseteq P$) is
because \emph{sureness} is a stronger requirement that
\emph{possibleness}.  A full description of the annotation guidelines
is available in a document available with the alignment software on
the first author's home page.  Here, we summarize the main points.

The most important instruction that annotators were given was to align
everything in the summary to \emph{something}.  This was not always
possible, as we will discuss shortly, but by and large it was an
appropriate heuristic.  The second major instruction was to choose
alignments with \emph{maximal consecutive length}: If there are two
possible alignments for a phrase, the annotators were instructed to
choose the one that will result in the longest consecutive alignment.
For example, in Figure~\ref{alignment-ex}, this rule governs the
choice of the alignment of the word ``Macintosh'' on the summary side:
lexically, it could be aligned to the final occurrence of the word
``Macintosh'' on the document side, but by aligning it to ``Apple
Macintosh systems,'' we are able to achieve a longer consecutive
sequence of aligned words.

The remainder of the instructions have to do primarily with clarifying
particular linguistic phenomenon, including punctuation, anaphora (for
entities, annotators are told to feel free to align names to pronouns,
for instance) and metonymy, null elements, genitives, appositives,
and ellipsis.

\subsection{Annotator agreement}

To compute annotator agreement, we employed the kappa statistic.  To
do so, we treat the problem as a sequence of binary decisions: given a
single summary word and document word, should the two be aligned.  To
account for phrase-to-phrase alignments, we first converted these into
word-to-word alignments using the ``all pairs'' heuristic.  By looking
at all such pairs, we wound up with $7.2$ million items over which to
compute the kappa statistic (with two annotators and two categories).
Annotator agreement was strong for sure alignments and fairly weak for
possible alignments.  When considering only sure alignments, the kappa
statistic for agreement was $0.63$ (though it dropped drastically to
$0.42$ on possible alignments).

In performing the annotation, we found that punctuation and
non-content words are often very difficult to align (despite the
discussion of these issues in the alignment guidelines).  The primary
difficulty with function words is that when the summarizers have
chosen to reorder words to use slightly different syntactic
structures, there are lexical changes that are hard to
predict\footnote{\setlength{\baselineskip}{1\baselineskip}For example, the change from ``I gave a gift to the
boy.'' to ``The boy received a gift from me.'' is relatively
straightforward; however, it is a matter of opinion whether ``to'' and
``from'' should be aligned -- they serve the same role, but certainly
do not mean the same thing.}.  Fortunately, for many summarization
tasks, it is much more important to get content words right, rather
than functions words.  When words on a stop list of $58$ function
words and punctuation were ignored, the kappa value rose to $0.68$.
Carletta \shortcite{carletta96kappa} has suggested that kappa values
over $0.80$ reflect very strong agreement and that kappa values
between $0.60$ and $0.80$ reflect good agreement.\footnote{\setlength{\baselineskip}{1\baselineskip}All
annotator agreement figures are calculated only on the last 40
\docabs\ pairs, which were annotated independently.}

\subsection{Results of annotation}

After the completion of these alignments, we can investigate some of
their properties.  Such an investigation is interesting both from the
perspective of designing a model, as well as from a linguistic
perspective.

In the alignments, we found that roughly $16\%$ of the \emph{abstract}
words are left unaligned.  This figure includes both standard lexical
words, as well as punctuation.  Of this $16\%$, $4\%$ are punctuation
marks (though not all punctuation is unaligned) and $7\%$ are function
words.  The remaining $5\%$ are words that would typically be
considered content words.  This rather surprising result tells us that
any model we build needs to be able to account for a reasonable
portion of the abstract to not have a direct correspondence to any
portion of the document.

To get a sense of the importance of producing alignments at the
\emph{phrase} level, we computed that roughly $75\%$ of the alignments
produced by humans involve only one word on both sides.  In $80\%$ of
the alignments, the summary side is a single word (thus in $5\%$ of
the cases, a single summary word is aligned to more than one document
word).  In $6.1\%$ of the alignments, the summary side involved a
phrase of length two, and in $2.2\%$ of the cases it involved a phrase
of length three.  In all these numbers, care should be taken to note
that the humans were instructed to produce phrase alignments
\emph{only} when word alignments were impossible.  Thus, it is
entirely likely that summary word $i$ is aligned to document word $j$
and summary word $i+1$ is aligned to document word $j+1$, in which
case we count this as two singleton alignments, rather than an
alignment of length two.  These numbers suggest that looking at
phrases in addition to words is empirically important.

Lexical choice is another important aspect of the alignment process.
Of all the aligned summary words and phrases, the corresponding
document word or phrase was exactly the same as that on the summary
side in $51\%$ of the cases.  When this constraint was weakened to
looking only at stems (for multi-word phrases, a match meant that each
corresponding word matched up to stem), this number rose to $67\%$.
When broken down into cases of singletons and non-singletons, we saw
that $86\%$ of singletons are identical up to stem and $48\%$ of
phrases are identical up to stem.  This suggests that looking at
stems, rather than lexical items, is useful.

\begin{figure}[!bt]
\mbox{\psfig{figure=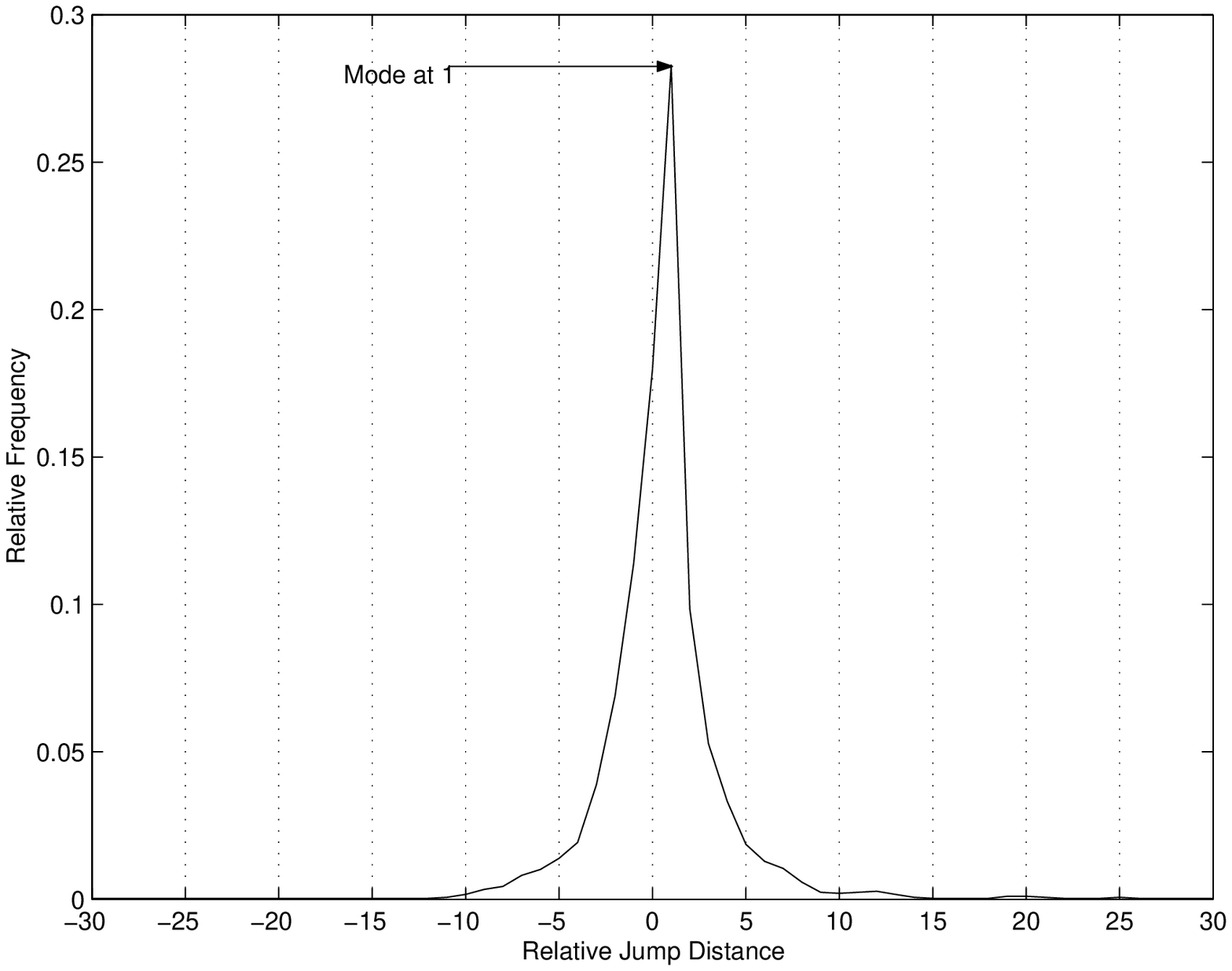,width=2.5in}}
\caption{\setlength{\baselineskip}{1\baselineskip}Analysis of the motion observed in documents when considering
  a movement of $+1$ on the summary side.}
\label{fig:movement-delta}
\end{figure}

Finally, we investigated the issue of adjacency in the alignments.
Specifically, we consider the following question: given that a summary
phrase ending at position $i$ is aligned to a document phrase ending
at position $j$, what is a likely position in the document for the
summary phrase beginning at position $i+1$.  It turns out that this is
overwhelmingly $j+1$.  In Figure~\ref{fig:movement-delta}, we have
plotted the frequencies of such relative ``jumps'' over the
human-aligned data.  This graph suggests that a model biased toward
stepping forward monotonically in the document is likely to be
appropriate.  However, it should also be noted that \emph{backward}
jumps are also quite common, suggesting that a monotonic alignment
model is inappropriate for this task.

\section{Statistical Alignment Model} \label{sec:model}

Based on linguistic observations from the previous section, we reach
several conclusions regarding the development of a statistical model
to produce such alignments.  First, the model should be able to
produce alignments between phrases of arbitrary length (but perhaps
with a bias toward single words).  Second, it should not be
constrained by any assumptions of monotonicity or word (or stem)
identity, but which might be able to realize that monotonicity and
word and stem identity are \emph{good indicators} of alignment.
Third, our model must be able to account for words on the abstract
side that have no correspondence on the document side (following the
terminology from the machine translation community, we will refer to
such words as ``null generated'').

\subsection{Generative story}

Based on these observations, and with an eye toward computational
tractability, we posit the following \emph{generative story} for how a
summary is produced, given a document:

\begin{enumerate}
\item Repeat until the whole summary is generated:
\begin{enumerate}
\item Choose a document position $j$ and ``jump'' there.
\item Choose a document phrase length $l$.
\item Generate a summary phrase based on the document phrase spanning
  positions $j$ to $j+l$.
\end{enumerate}
\item ``Jump'' to the end of the document.
\end{enumerate}

In order to account for ``null generated'' summary words, we augment
the above generative story with the option to jump to a specifically
designated ``null state'' from which a summary phrase may be generated
without any correspondence in the document.  From inspection of the
human-aligned data, most such null generated words are function words
or punctuation; however, in some cases, there are pieces of
information in the summary that truly did not exist in the original
document.  The null generated words can account for these as well
(additionally, the null generated words allow the model to ``give up''
when it cannot do anything better).  We require that summary phrases
produced from the null state have length 1, so that in order to
generate multiple null generated words, they must be generated
independently.

\begin{figure}[!bt]
\mbox{\psfig{figure=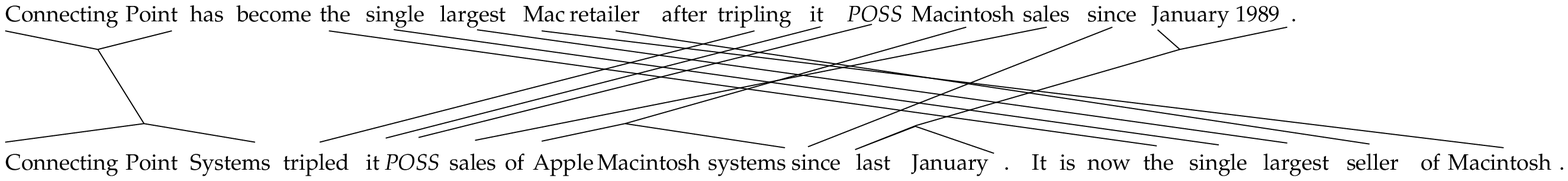,width=33pc}}
\small
\begin{itemize}
\vspace{-1mm}\item[-] Jump to the first document word and choose a length of $3$.
\vspace{-1mm}\item[-] Generate the summary phrase ``Connecting Point'' based on the
  document phrase ``Connecting Point Systems.''
\vspace{-1mm}\item[-] Jump to a null state.
\vspace{-1mm}\item[-] Generate the summary word ``has'' from null.
\vspace{-1mm}\item[-] Generate the summary word ``become'' from null.
\vspace{-1mm}\item[-] Jump our of ``null'' to the fourth document word and choose a length of $1$.
\vspace{-1mm}\item[-] Generate the summary phrase ``tripling'' given ``tripled.''
\vspace{-1mm}\item[-] Jump to the fifth document word and choose a length of $1$.
\vspace{-1mm}\item[] \dots
\vspace{-1mm}\item[-] Jump to the thirteenth document word (``last'') and choose a
  length of $2$.
\vspace{-1mm}\item[-] Generate the summary phrase ``January 1989'' given ``last
  January.''
\vspace{-1mm}\item[-] Jump to a null state.
\vspace{-1mm}\item[-] Generate the summary word ``.'' from null.
\vspace{-1mm}\item[-] Jump out of null to the end of the document.
\end{itemize}
\caption{\setlength{\baselineskip}{1\baselineskip}Beginning and end of the generative process that gave rise to
  the alignment in Figure~\ref{alignment-ex}, which is reproduced here
  for convenience.}
\label{fig:genstory}
\end{figure}

In Figure~\ref{fig:genstory}, we have shown a portion of the
generative process that would give rise to the alignment in
Figure~\ref{alignment-ex}.

This generative story implicitly induces an alignment between the
document and the summary: the summary phrase is considered to be
aligned to the document phrase that ``generated'' it.  In order to
make this computationally tractable, we must introduce some
conditional independence assumptions.  Specifically, we assume the
following:

\begin{enumerate}
\item Decision (a) in our generative story depends only on the
  position of the end of the current document phrase (i.e., $j+l$).
\item Decision (b) is conditionally independent of every other
  decision.
\item Decision (c) depends only on the phrase at the current document
  position.
\end{enumerate}

\subsection{Statistical model}

Based on the generative story and independence assumptions described
above, we can model the entire summary generation process according to
two distributions:

\begin{itemize}
\item $\jump(j' \| j+l)$, the probability of jumping to
  position $j'$ in the document when the previous phrase ended at
  position $j+l$.

\item $\rewrite(\vec s \| \vec d_{j:j+l})$, the ``rewrite'' probability of
  generating summary phrase $\vec s$ given that we are considering the
  sub-phrase of $\vec d$ beginning at position $j$ and ending at
  position $j+l$.
\end{itemize}

Specific parameterizations of the distributions $\jump$ and $\rewrite$
will be discussed in Section~\ref{sec:parameterization} to enable the
focus here to be on the more general problems of inference and
decoding in such a model.  The model described by these independence
assumptions very much resembles that of a hidden Markov model (HMM),
where states in the state space are document ranges and emissions are
summary words.  The difference is that instead of generating a single
word in each transition between states, an entire phrase is generated.
This difference is captured by the \emph{semi-Markov model} or
\emph{segmental HMM} framework, described in great detail by
Ostendorf, Digalakis, and Kimball \shortcite{ostendorf96segment}; see
also
\cite{ferguson80variable,gales93segmental,mitchell95duration,smyth97pins,ge00segmental,aydin04protein}
for more detailed descriptions of these models as well as other
applications in speech processing and computational biology.  In the
following subsections, we will briefly discuss the aspects of
inference that are relevant to our problem, but the interested reader
is directed to \cite{ostendorf96segment} for more details.

\subsection{Creating the state space}

Given our generative story, we can construct a semi-HMM to calculate
precisely the alignment probabilities specified by our model in an
efficient manner.  A semi-HMM is fully defined by a state space (with
designated start and end states), an output alphabet, transition
probabilities and observations probabilities.  The semi-HMM functions
like an HMM: beginning at the start state, stochastic transitions are
made through the state space according to the transition
probabilities.  At each step, one or more observations is generated.
The machine stops when it reaches the end state.

\begin{figure}[t]
\mbox{\psfig{figure=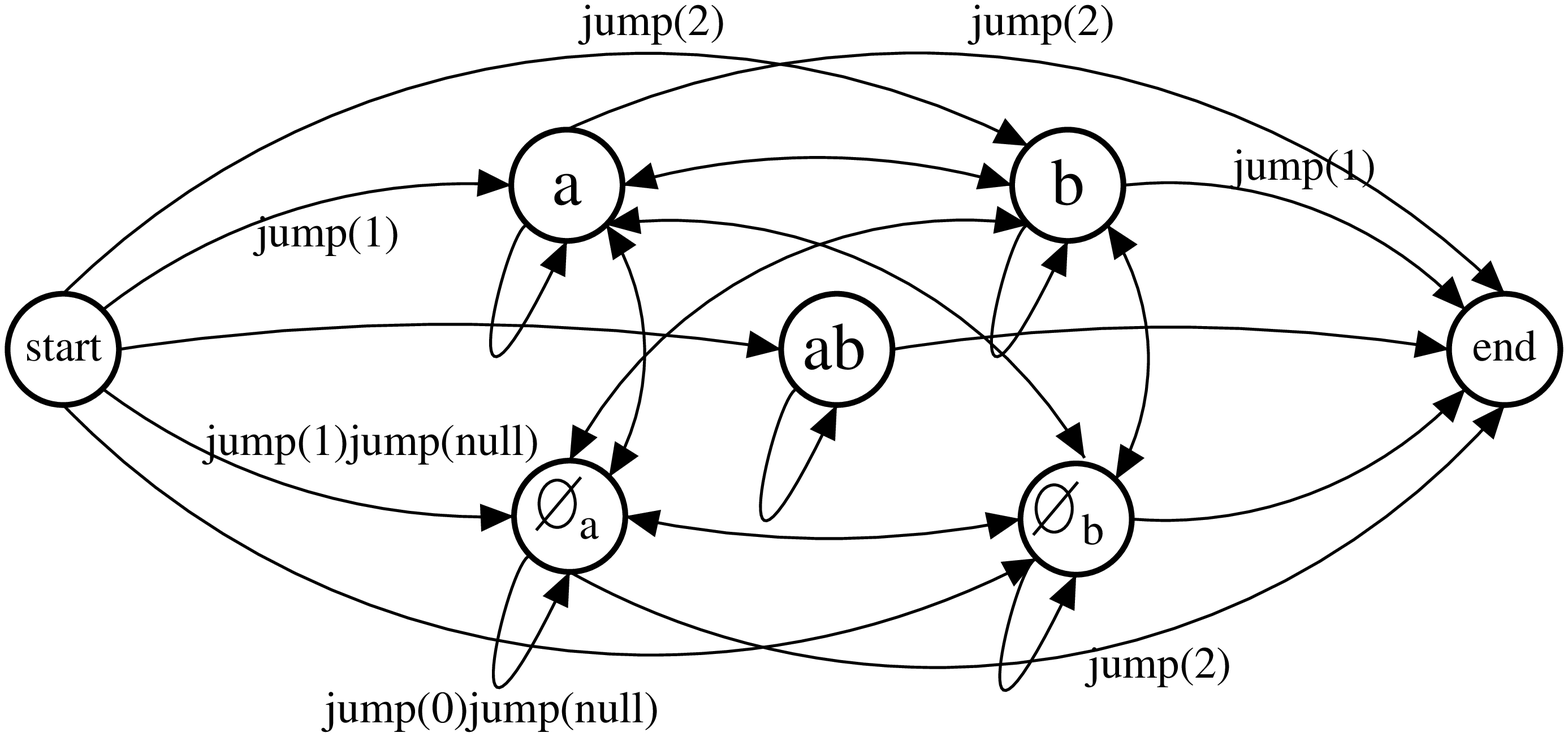,width=4.5in}}
\caption{\setlength{\baselineskip}{1\baselineskip}Schematic drawing of the semi-HMM (with some transition probabilities) for the document ``a b''}
\label{pbhmm-ex}
\end{figure}

In our case, the state set is large, but well structured.  There is a
unique initial state $\langle\text{start}\rangle$, a unique final
state $\langle\text{end}\rangle$, and a state for each possible
document phrase.  That is, for a document of length $n$, for all $1
\leq i \leq i' \leq n$, there is a state that corresponds to the
document phrase beginning at position $i$ and ending at position $i'$,
which we will refer to as $r_{i,i'}$.  There is also a null state for
each document position $r_{\emptyset,i}$.  Thus, $S = \{
\langle\text{start}\rangle, \langle\text{end}\rangle \} \cup \{
r_{i,i'} : 1 \leq i \leq i' \leq n \} \cup \{ r_{\emptyset,i} : 1 \leq
i \leq n \}$.  The output alphabet consists of each word found in $S$,
plus the end-of-sentence word $\omega$.  We only allow the word
$\omega$ to be emitted on a transition to the end state.  The
transition probabilities are managed by the jump model, and the
emission probabilities are managed by the rewrite model.

Consider the document ``a b'' (the semi-HMM for which is shown in
Figure~\ref{pbhmm-ex}) in the case when the corresponding summary is
``c d.''  Suppose the correct alignment is that ``c d'' is aligned to
``a'' and ``b'' is left unaligned.  Then, the path taken through the
semi-HMM is $\langle\text{start}\rangle \fto a \fto
\langle\text{end}\rangle$.  During the transition
$\langle\text{start}\rangle \fto a$, ``c d'' is emitted.  During the
transition $a \fto \langle\text{end}\rangle$, $\omega$ is emitted.

\subsection{Expectation maximization}

The alignment task, as described above, is a chicken and egg problem:
if we \emph{knew} the model components (namely, the rewrite and jump
tables), we would be able to efficiently find the best alignment.
Similarly, if we knew the correct alignments, we would be able to
estimate the model components.  Unfortunately, we have neither.
Expectation maximization is a general technique for learning in such
chicken and egg problems \cite{dempster77em,boyles83em,wu83em}.  The
basic idea is to make a \emph{guess} at the alignments, then use this
guess to estimate the parameters for the relevant distributions.  We
can use these re-estimated distributions to make a better guess at the
alignments, and then use these (hopefully better) alignments to
re-estimate the parameters.

Formally, the EM family of algorithms tightly bound the $\log$ of an
expectation of a function by the expectation of $\log$ of that
function, through the use of Jensen's inequality
\cite{jensen06inequality}.  The tightness of the bound means that when
we attempt to estimate the model parameters, we may do so over
\emph{expected} alignments, rather that the true (but unknown)
alignments.  EM gives formal guarantees of convergence, but is only
guaranteed to find local maxima.

\subsection{Model inference}

\begin{figure}[t]
\center \small
\begin{eqnarray}
\al_j(t) &=& \p{\vec s_{1:t-1}, \textnormal{doc posn} = j}
        = \sum_{t'=0}^{t-1}
        \sum_{i\in S}
          \al_{i}(t'+1)
          \jump(j \| i)
          \rewrite(\vec s_{1:t-1} \| d_j)
          \nonumber\\
\be_i(t) &=& \p{\vec s_{t:T} \| \textnormal{doc posn} = i} 
        = \sum_{t'=t}^T
        \sum_{j\in S}
            \jump(j \| i)
            \rewrite(\vec s_{t:t'} \| d_j)
            \be_j(t'+1)
          \nonumber\\
\zeta_j(t) &=& \max_{i,t'}
                 \zeta_i(t') \jump(j \| i) \rewrite(\vec s_{t':t-1} \| d_j) \nonumber\\
\tau_{i,j}(t',t) &=& \Ep\{\textnormal{\# transitions $i \leadsto j$
    emitting $s_{t':t}$}\} 
=
 \frac {\al_i(t')
          \jump(j \| i)
          \rewrite(\vec s_{t':t} \| d_j)
           \be_j(t+1)}
          {\p{\vec s_{1:T}}} \nonumber\\
\hat a_{i,j} &=& 
  \frac {\Ep\{\textnormal{\#  transitions $i \leadsto j$}\}}
        {\Ep\{\textnormal{\# transitions $i \leadsto ?$}\}}
  =
  \frac {\sum_{t'=1}^T \sum_{t=t'}^T \tau_{i,j}(t',t)}
        {\sum_{t'=1}^T \sum_{t=t'}^T \sum_{j' \in S} \tau_{i,j'}(t',t)} \nonumber\\
\hat b_{i,j,\vec k} &=&
  \frac {\Ep\{\textnormal{\# transitions $i \leadsto j$ with $\vec w$ observed}\}}
        {\Ep\{\textnormal{\# transitions $i \leadsto j$}\}} 
  =
  \frac {\sum_{t : s_{t:t+|\vec k|-1} = \vec k}
               \tau_{i,j}(t,t+|\vec k|-1)}
        {\sum_{t'=1}^T \sum_{t=t'}^T \tau_{i,j}(t',t)} \nonumber
\end{eqnarray}
\caption{\setlength{\baselineskip}{1\baselineskip}Summary of inference equations for a semi-Markov model}
\label{fig:semi-eqn}
\end{figure}

All the inference techniques utilized in this paper are standard
applications of semi-Markov model techniques.  The relevant equations
are summarized in Figure~\ref{fig:semi-eqn} and described below.  In
all these equations, the variables $t$ and $t'$ range over phrases in
the summary (specifically, the phrase $\vec s_{t:t'}$), and the
variables $i$ and $j$ range over phrases in the document.  The
interested reader is directed to \cite{ostendorf96segment} for more
details on the generic form of these models and their inference
techniques.

The standard, and simplest inference problem in our model is to
compute $p(\vec s \| \vec d)$ by summing over all possible alignments.
A na\"ive implementation of this will take time exponential in the
length of the summary.  Instead, we are able to employ a variant of
the \emph{forward} algorithm to compute these probabilities
recursively.  The basic idea is to compute the probability of
generating a prefix of the summary and ending up at a particular
position in the document (this is known as the forward probability).
Since our independence assumptions tell us that it does not matter how
we got to this position, we can use this forward probability to
compute the probability of taking one more step in the summary.  At
the end, the desired probability $p(\vec s \| \vec d)$ is simply the
forward probability of reaching the end of the summary and document
simultaneously.  The forward probabilities are calculated in the
``$\al$ table'' in Figure~\ref{fig:semi-eqn}.  This equation
essentially says that the probability of emitting the first $t-1$ words
of the summary and ending at position $j$ in the document can be
computed by summing over our previous position ($t'$) and previous
state ($i$) and multiplying the probability of getting there
($\al_i(t'+1)$) with the probability of moving from there to the
current position.

The second standard inference problem is the calculation of the best
alignment: the Viterbi alignment.  This alignment can be computed in
exactly the same fashion as the forward algorithm, with two small
changes.  First, the forward probabilities implicitly include a sum
over all previous states, whereas the Viterbi probabilities replace
this with a max operator.  Second, in order to recover the actual
Viterbi alignment, we keep track of which previous state this max
operator chose.  This is computed by filling out the $\ze$ table from
Figure~\ref{fig:semi-eqn}.  This is almost identical to the
computation of the forward probabilities, except that instead of
summing over all possible $t'$ and $i$, we take the maximum over those
variables.

The final inference problem is parameter re-estimation.  In the case
of standard HMMs, this is known as the Baum-Welch or Baum-Eagon or
Forward-Backward algorithm \cite{baum66hmms,baum67hmms}.  By
introducing ``backward'' probabilities, analogous to the forward
probabilities, we can compute alignment probabilities of suffixes of
the summary.  The backward table is the ``$\be$ table'' in
Figure~\ref{fig:semi-eqn}, which is analogous to the $\al$ table,
except that the computation proceeds from the end to the start.

By combining the forward and backward probabilities, we can compute
the expected number of times a particular alignment was made (the
E-step in the EM framework).  Based on these expectations, we can
simply sum and normalize to get new parameters (the M-step).  The
expected transitions are computed according to the $\tau$ table, which
makes use of the forward and backward probabilities.  Finally, the
re-estimated jump probabilities are given by $\hat a$ and the
re-estimated rewrite probabilities are given by $\hat b$, which are
essentially relative frequencies of the fractional counts given by the
$\tau$s.

The computational complexity for the Viterbi algorithm and for the
parameter re-estimation is $\O{N^2T^2}$, where $N$ is the length of
the summary, and $T$ is the number of states (in our case, $T$ is
roughly the length of the document times the maximum phrase length
allowed).  However, we will typically bound the maximum length of a
phrase; otherwise we are unlikely to encounter enough training data to
get reasonable estimates of emission probabilities.  If we enforce a
maximum observation sequence length of $l$, then this drops to
$\O{N^2Tl}$.  Moreover, if the transition network is sparse, as it is
in our case, and the maximum out-degree of any node is $b$, then the
complexity drops to $\O{NTbl}$.

\section{Model Parameterization} \label{sec:parameterization}

Beyond the conditional independence assumptions made by the semi-HMM,
there are nearly no additional constraints that are imposed on the
parameterization (in term of the $\jump$ and $\rewrite$ distributions)
of the model.  There is one additional technical requirement involving
parameter re-estimation, which essentially says that the expectations
calculated during the forward-backward algorithm must be sufficient
statistics for the parameters of the $\jump$ and $\rewrite$ models.
This constraint simply requires that whatever information we need to
re-estimate their parameters is available to us from the
forward-backward algorithm.

\subsection{Parameterizing the jump model}

Recall that the responsibility of the jump model is to compute
probabilities of the form $\jump(j' \| j)$, where $j'$ is a new
position and $j$ is an old position.  We have explored several
possible parameterizations of the jump table.  The first simply
computes a table of likely jump distances (i.e., jump forward 1, jump
backward 3, etc.).  The second models this distribution as a Gaussian
(though, based on Figure~\ref{fig:movement-delta} this is perhaps not
the best model).  Both of these models have been explored in the
machine translation community.  Our third parameterization employs a
novel \emph{syntax-aware} jump model that attempts to take advantage
of local syntactic information in computing jumps.

\subsubsection{The relative jump model.}

\begin{table}[tb]
\tcaption{\setlength{\baselineskip}{1\baselineskip}Jump probability decomposition; the source state is either
  the designated start state, the designated end state, a document
  phrase position spanning from $i$ to $i'$ (denoted $r_{i,i'}$) for a
  null state corresponding to position $i$ (denoted
  $r_{\emptyset,i}$).}
\label{jumps}
\begin{tabular}{|l|l|l|}
\hline
source & target & probability \\
\hline
$\langle\text{start}\rangle$ & $r_{i,i'}$ & $\jump_\textrm{rel}(i)$ \\
$r_{i,i'}$ & $r_{j,j'}$ & $\jump_\textrm{rel}(j-i')$ \\
$r_{i,j'}$ & $\langle\text{end}\rangle$ & $\jump_\textrm{rel}(m+1-i')$ \\
\hline
$\langle\text{start}\rangle$ & $r_{\emptyset,i}$ & $\jump_\textrm{rel}(\emptyset) \jump_\textrm{rel}(i)$ \\
$r_{\emptyset,i}$ & $r_{j,j'}$ & $\jump_\textrm{rel}(j-i)$ \\
$r_{\emptyset,i}$ & $r_{\emptyset,j}$ & $\jump_\textrm{rel}(\emptyset) \jump_\textrm{rel}(j-i)$ \\
$r_{\emptyset,i}$ & $\langle\text{end}\rangle$ & $\jump_\textrm{rel}(m+1-i)$\\
$r_{i,i'}$ & $r_{\emptyset,j}$ & $\jump_\textrm{rel}(\emptyset) \jump_\textrm{rel}(j-i')$ \\
\hline
\end{tabular}
\end{table}

In the relative jump model, we keep a table of counts for each
possible jump distance, and compute $\jump(j' \| j) =
\jump_\textrm{rel}(j'-j)$.  Each possible jump type and its associated
probability is shown in Table~\ref{jumps}.  By these calculations,
regardless of document phrase lengths, transitioning forward between
two consecutive segments will result in $\jump_\textrm{rel}(1)$.  When
transitioning from the start state $p$ to state $r_{i,i'}$, the value
we use a jump length of $i$.  Thus, if we begin at the first word in
the document, we incur a transition probability of $j{1}$.  There are
no transitions into $p$.  We additionally remember a specific
transition $\jump_\textrm{rel}(\emptyset)$ for the probability of
transitioning to a null state.  It is straightforward to estimate
these parameters based on the estimations from the forward-backward
algorithm.  In particular, $\jump_\textrm{rel}(i)$ is simply the
relative frequency of length $i$ jumps and
$\jump_\textrm{rel}(\emptyset)$ is simply the count of jumps that end
in a null state to the total number of jumps.  The null state
``remembers'' the position we ended in before we jumped there and so
to jump \emph{out} of a null state, we make a jump based on this
previous position.\footnote{\setlength{\baselineskip}{1\baselineskip}In order for the null state to remember
where we were, we actually introduce one null state for each document
position, and require that from a document phrase $\vec d_{i:j}$, we
can only jump to null state $\emptyset_j$.}

\subsubsection{Gaussian jump model.}

The Gaussian jump model attempts to alleviate the sparsity of data
problem in the relative jump model by assuming a parametric form to
the jumps.  In particular, we assume there is a mean jump length $\mu$
and a jump variance $\si^2$, and then the probability of a jump of
length $i$ is given by:

\begin{equation}
i \by \Nor(\mu, \si^2) \varpropto 
  \exp\left[\frac 1 {\si^2} (i - \mu)^2\right]
\end{equation}

Some care must be taken in employing this model, since the normal
distribution is defined over a continuous space.  Thus, when we
discretize the calculation, the normalizing constant changes slightly
from that of a continuous normal distribution.  In practice, we
normalize by summing over a sufficiently large range of possible $i$s.
The parameters $\mu$ and $\si^2$ are estimated by computing the mean
jump length in the expectations and its empirical variance.  We model
null states identically to the relative jump model.

\subsubsection{Syntax-aware jump model.}

Both of the previously described jump models are extremely na\"ive, in
that they look only at the distance jumped, and completely ignore what
is being jumped over.  In the syntax-aware jump model, we wish to
enable the model to take advantage of syntactic knowledge in a very
weak fashion.  This is quite different from the various approaches to
incorporating syntactic knowledge into machine translation systems,
wherein strong assumptions about the possible syntactic operations are
made \cite{yamada01,eisner03treemappings,gildea03alignment}.

\begin{figure}[t]
\mbox{\psfig{figure=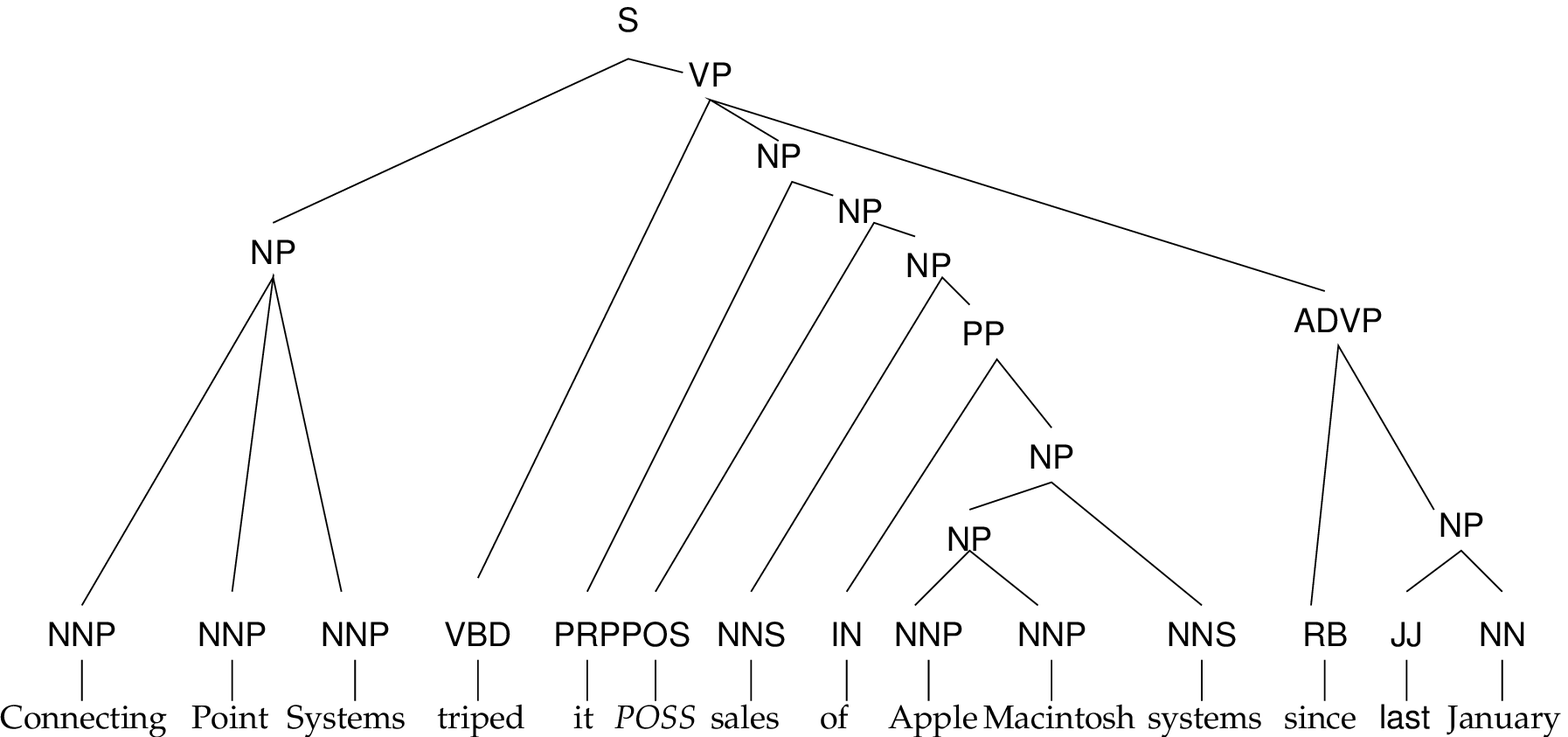,width=4.5in}}
\caption{\setlength{\baselineskip}{1\baselineskip}The syntactic tree for an example document sentence.}
\label{fig:syntax-tree}
\end{figure}

To motivate this model, consider the first document sentence shown
with its syntactic parse tree in Figure~\ref{fig:syntax-tree}.  Though
it is not always the case, forward jumps of distance more than one are
often indicative of skipped words.  From the standpoint of the
relative jump models, jumping over the four words ``tripled it~'s
sales'' and jumping over the four words ``of Apple Macintosh systems''
are exactly the same.  However, intuitively, we would be much more
willing to jump over the latter than the former.  The latter phrase is
a full syntactic constituent, while the first phrase is just a
collection of nearby words.  Furthermore, the latter phrase is a
prepositional phrase (and prepositional phrases might be more likely
dropped than other phrases) while the former phrase includes a verb, a
pronoun, a possessive marker and a plain noun.

To formally capture this notion, we parameterize the syntax-aware jump
model according to the types of phrases being jumped over.  That is,
to jump over ``tripled it~'s sales'' would have probability
$\jump_\textrm{syn}(\textsf{VBD PRP POS NNS})$ while to jump over ``of Apple Macintosh
systems'' would have probability $\jump_\textsf{syn}(\textsf{PP})$.  In order to
compute the probabilities for jumps over many components, we factorize
so that the first probability becomes $\jump_\textsf{syn}(\textsf{VBD}) \jump_\textsf{syn}(\textsf{PRP})
\jump_\textsf{syn}(\textsf{POS}) \jump_\textsf{syn}(\textsf{NNS})$.  This factorization explicitly
encodes our preference for jumping over single units, rather than
several syntactically unrelated units.

In order to work with this model, we must first parse the document
side of the corpus; we used Charniak's parser \cite{charniak97}.
Given the document parse trees, the re-estimation of the components of
this probability distribution is done by simply counting what sorts of
phrases are being jumped over.  Again, we keep a single parameter
$\jump_\textrm{syn}(\emptyset)$ for jumping to null states.  To handle
backward jumps, we simply consider a duplication of the tag set, where
$\jump_\textrm{syn}(\textsf{NP-f})$ denotes a forward jump over an
\textsf{NP} and $\jump_\textrm{syn}(\textsf{NP-b})$ denotes a backward
jump over an \textsf{NP}.\footnote{\setlength{\baselineskip}{1\baselineskip}In general, there are many ways to
get from one position to another.  For instance, to get from
``systems'' to ``January'', we could either jump forward over an
\textsf{RB} and a \textsf{JJ}, or we could jump forward over an
\textsf{ADVP} and backward over an \textsf{NN}.  In our version, we
restrict that all jumps are in the same direction, and take the
shortest jump sequence, in terms of number of nodes jumped over.}

\subsection{Parameterizing the rewrite model}

As observed from the human-aligned summaries, a good rewrite model
should be able to account for alignments between identical word and
phrases, between words that are identical up to stem, and between
different words.  Intuition (as well as further investigations of the
data) also suggest that \emph{synonymy} is an important factor to take
into consideration in a successful rewrite model.  We account for each
of these four factors in four separate components of the model and
then take a linear interpolation of them to produce the final
probability:

\begin{align}
\rewrite(\vec s \| \vec d) &=
\la_{\textrm{id}}   \rewrite_{\textrm{id}  }(\vec s \| \vec d) +
\la_{\textrm{stem}} \rewrite_{\textrm{stem}}(\vec s \| \vec d) \\
&+
\la_\textrm{wn}     \rewrite_{\textrm{wn}  }(\vec s \| \vec d) +
\la_{\textrm{rw}}   \rewrite_{\textrm{rw}  }(\vec s \| \vec d)
\end{align}

where the $\la$s are constrained to sum to unity.  The four rewrite
distributions used are: $\textrm{id}$ is a word identity model, which
favors alignment of identical words; $\textrm{stem}$ is a model
designed to capture the notion that matches at the stem level are
often sufficient for alignment (i.e., ``walk'' and ``walked'' are
likely to be aligned); $\textrm{wn}$ is a rewrite model based on
similarity according to WordNet; and $\textrm{wr}$ is the basic
rewrite model, similar to a translation table in machine translation.
These four models are described in detail below, followed by a
description of how to compute their $\la$s during EM.

\subsubsection{Word identity rewrite model.}

The form of the word identity rewrite model is: $\rewrite_{\textrm{id}}(\vec
s \| \vec d) = \de_{\vec s = \vec d}$.  That is, the probability is
$1$ exactly when $\vec s$ and $\vec d$ are identical, and $0$ when
they differ.  This model has no parameters.

\subsubsection{Stem identity rewrite model.}

The form of the stem identity rewrite model is very similar to that of
the word identity model:

\begin{equation}
\rewrite_{\textrm{stem}}(\vec s \| \vec d) = 
  \frac 1 {Z_{\vec d}}
  \de_{|\vec s| = |\vec d|}
  \prod_{i=1}^{|\vec s|}
    \de_{\textit{stem}(s_i) = \textit{stem}(d_i)}
\end{equation}

That is, the probability of a phrase $\vec s$ given $\vec d$ is
uniform over all phrases $\vec s'$ that match $\vec d$ up to stem (and
are of the same length, i.e., $|\vec s'| = |\vec d|$), and zero
otherwise.  The normalization constant is computed offline based on a
pre-computed vocabulary.  This model also has no parameters.

\subsubsection{WordNet rewrite model.}

In order to account for synonymy, we allow document phrases to be
rewritten to semantically ``related'' summary phrases.  To compute the
value for $\rewrite_\textrm{wn}(\vec s \| \vec d)$ we first require
that both $\vec s$ and $\vec d$ can be found in WordNet.  If either
cannot be found, then the probability is zero.  If they both can be
found, then the graph distance between their first senses is computed
(we traverse the hypernymy tree up until they meet).  If the two paths
do not meet, then the probability is again taken to be zero.  We place
an exponential model on the hypernym tree-based distance:

\begin{equation}
\rewrite_\textrm{wn}(\vec s \| \vec d) = \frac 1 {Z_{\vec d}} \exp \left[ -
  \eta \textit{dist}(\vec s, \vec d) \right]
\end{equation}

Here, $\textit{dist}$ is calculated distance, taken to be $+\infty$
whenever either of the failure conditions is met.  The single
parameter of this model is $\eta$, which is computed according to the
maximum likelihood criterion from the expectations during training.
The normalization constant $Z_{\vec d}$ is calculated by summing over
the exponential distribution for all $\vec s'$ that occur on the
summary side of our corpus.

\subsubsection{Lexical rewrite model.}

The lexical rewrite model is the ``catch all'' model to handle the
cases not handled by the above models.  It is analogous to a
translation-table (t-table) in statistical machine translation (we
will continue to use this terminology for the remainder of the
article), and simply computes a matrix of (fractional) counts
corresponding to all possible phrase pairs.  Upon normalization, this
matrix gives the rewrite distribution.

\subsubsection{Estimation of the weight parameters.}

In order to weight the four models, we need to estimate values for the
$\la$ components.  This computation can be performed inside of the EM
iterations, by considering, for each rewritten pair, \emph{its}
expectation of belonging to each of the models.  We use these
expectations to maximize the likelihood with respect to the $\la$s and
then normalize them so they sum to one.

\subsection{Model priors}

In the standard HMM case, the learning task is simply one of parameter
estimation, wherein the maximum likelihood criterion under which the
parameters are typically trained performs well.  However, in our
model, we are, in a sense, simultaneously estimating parameters and
selecting a \emph{model}: the model selection is taking place at the
level of deciding how to segment the observed summary.  Unfortunately,
in such model selection problems, likelihood increases monotonically
with model complexity.  Thus, EM will find for us the most complex
model; in our case, this will correspond to a model in which the
entire summary is produced at once, and no generalization will be
possible.

This suggests that a criterion \emph{other than} maximum likelihood
(ML) is more appropriate.  We advocate the \emph{maximum a posteriori}
(MAP) criterion in this case.  While ML optimizes the probability of
the data given the parameters (the likelihood), MAP optimizes the
product of the probability of the parameters with the likelihood (the
unnormalized posterior).  The difficulty in our model that makes ML
estimation perform poorly is centered in the lexical rewrite model.
Under ML estimation, we will simply insert an entry in the t-table for
the entire summary for some uncommon or unique document word and be
done.  However, \emph{a priori} we do not believe that such a
parameter is likely.  The question then becomes how to express this in
a way that inference remains tractable.

From a statistical point of view, the t-table is nothing but a large
multinomial model (technically, one multinomial for each possible
document phrase).  Under a multinomial distribution with parameter
$\vec \th$ with $J$-many components (with all $\th_j$ positive and
summing to one), the probability of an observation $\vec x$ is given
by (here, we consider $\vec x$ to be a vector of length $J$ in which
all components are zero except for one, corresponding to the actual
observation): $\p{\vec x \| \vec \th} = \prod_{j=1}^J \th_j^{x_j}$.

This distribution belongs to the \emph{exponential family} and
therefore has a natural \emph{conjugate} distribution.  Informally,
two distributions are conjugate if you can multiply them together and
get the original distribution back.  In the case of the multinomial,
the conjugate distribution is the Dirichlet distribution.  A Dirichlet
distribution is parameterized by a vector $\vec \al$ of length $J$
with $\al_j \geq 0$, but not necessarily summing to one.  The
Dirichlet distribution can be used as a prior distribution over
multinomial parameters and has density: $\p{\vec \th \| \vec \al} =
\frac {\Ga\left(\sum_{j=1}^J \al_j\right)} {\prod_{j=1}^J \Ga(\al_j)}
\prod_{j=1}^J \th_j^{\al_j - 1}$.  The fraction before the product is
simply a normalization term that ensures that the integral over all
possible $\vec \th$ integrates to one.

The Dirichlet is conjugate to the multinomial because when we compute
the posterior of $\vec \th$ given $\vec \al$ and $\vec x$, we arrive
back at a Dirichlet distribution: $\p{\vec \th \| \vec x, \vec \al}
\varpropto \p{\vec x \| \vec \th} \p{\vec \th \| \vec \al} \varpropto
\prod_{j=1}^J \th_j^{x_j + \al_j - 1}$.  This distribution has the
same density as the original model, but a ``fake count'' of $\al_j-1$
has been added to component $j$.  This means that \emph{if} we are
able to express our prior beliefs about the multinomial parameters
found in the t-table in the form of a Dirichlet distribution, the
computation of the MAP solution can be performed exactly as described
before, but with the appropriate fake counts added to the observed
variables (in our case, the observed variables are the alignments
between a document phrase and a summary phrase).  The application of
Dirichlet priors to standard HMMs has previously been considered in
signal processing \cite{gauvain94maximum}.  These fake counts act as a
smoothing parameter, similar to Laplace smoothing (Laplace smoothing
is the special case where $\al_j = 2$ for all $j$).

In our case, we believe that singleton rewrites are worth $2$ fake
counts, that lexical identity rewrites are worth $4$ fake counts and
that stem identity rewrites are worth $3$ fake counts.  Indeed, since
a singleton alignment between identical words satisfies \emph{all} of
these criterion, it will receive a fake count of $9$.  The selection
of these counts is intuitive, but, clearly, arbitrary.  However, this
selection was not ``tuned'' to the data to get better performance.  As
we will discuss later, inference in this model over the sizes of
documents and summaries we consider is quite computationally
expensive.  We appropriately specified this prior according to our
prior beliefs, and left the rest to the inference mechanism.

\subsection{Parameter initialization}

We initialize all the parameters uniformly, but in the case of the
rewrite parameters, since there is a prior on them, they are
effectively initialized to the maximum likelihood solution under their
prior.

\section{Experimental Results} \label{sec:results}

The experiments we perform are on the same Ziff-Davis corpus described
in the introduction.  In order to judge the quality of the alignments
produced, we compare them against the gold standard references
annotated by the humans.  The standard precision and recall metrics
used in information retrieval are modified slightly to deal with the
``sure'' and ``possible'' alignments created during the annotation
process.  Given the set $S$ of sure alignments, the set $S \subseteq
P$ of possible alignments and a set $A$ of hypothesized alignments we
compute the precision as $|A \cap P| / |A|$ and the recall as $|A \cap
S| / |S|$.

One problem with these definitions is that phrase-based models are
fond of making phrases.  That is, when given an abstract containing
``the man'' and a document also containing ``the man,'' a human will
align ``the'' to ``the'' and ``man'' to ``man.''  However, a
phrase-based model will almost always prefer to align the entire
phrase ``the man'' to ``the man.''  This is because it results in
fewer probabilities being multiplied together.

To compensate for this, we define soft precision (SoftP in the tables)
by counting alignments where ``a b'' is aligned to ``a b'' the same as
ones in which ``a'' is aligned to ``a'' and ``b'' is aligned to ``b.''
Note, however, that this is not the same as ``a'' aligned to ``a b''
and ``b'' aligned to ``b''.  This latter alignment will, of course,
incur a precision error.  The soft precision metric induces a new,
soft F-Score, labeled SoftF.

Often, even humans find it difficult to align function words and
punctuation.  A list of $58$ function words and punctuation marks
which appeared in the corpus (henceforth called the
\emph{ignore-list}) was assembled.  We computed precision and recall
scores both on all words, as well as on all words that do not appear
in the ignore-list.

\subsection{Systems compared}

Overall, we compare various parameter settings of our model against
three other systems.  First, we compare against two alignment models
developed in the context of machine translation.  Second, we compare
against the Cut and Paste model developed in the context of ``summary
decomposition'' by Jing \shortcite{jing:cl}.  Each of these systems
will be discussed in more detail shortly.  However, the machine
translation alignment models assume \emph{sentence pairs} as input.
Moreover, even though the semi-Markov model is based on efficient
dynamic programming techniques, it is still too inefficient to run on
very long \docabs\ pairs.

\begin{figure}[t]
\mbox{\psfig{figure=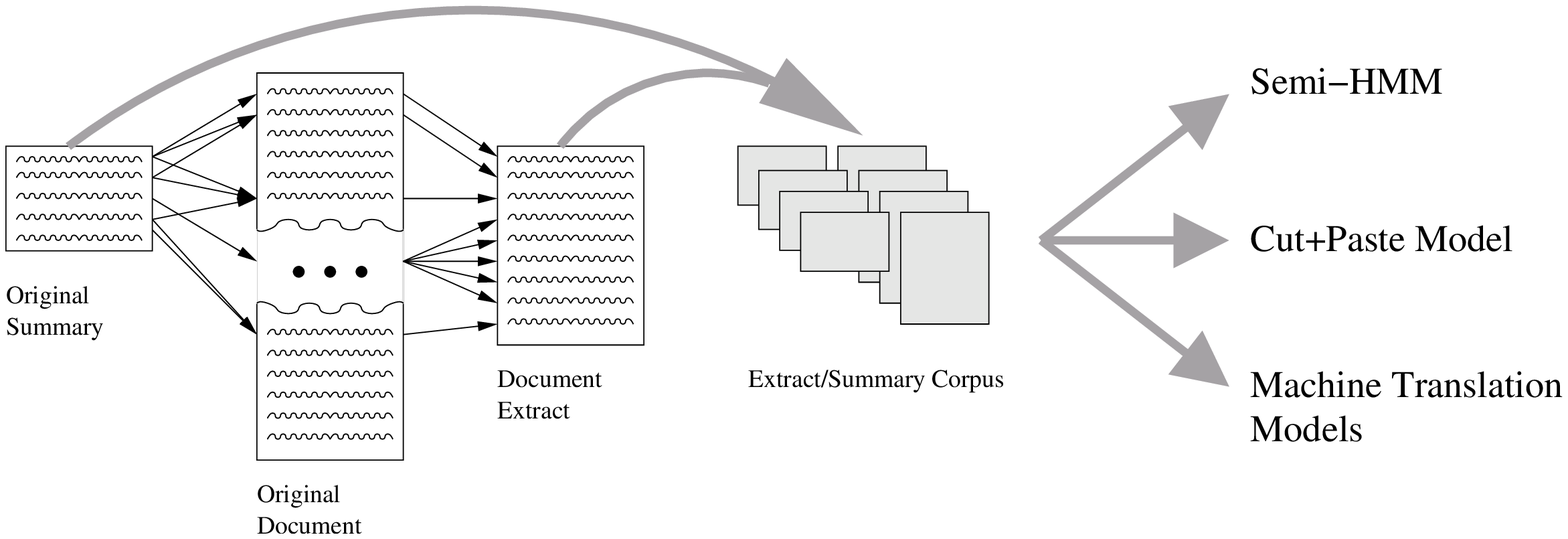,width=5in}}
\caption{\setlength{\baselineskip}{1\baselineskip}Pictorial representation of the conversion of the
  \docabs\ corpus to an \extabs\ corpus.}
\label{fig:doc2ext}
\end{figure}

To alleviate both of these problems, we preprocess our
\docabs\ corpus \emph{down to} an \extabs\ corpus,
and then subsequently apply our models to this smaller corpus (see
Figure~\ref{fig:doc2ext}).  In our data, doing so does not introduce
significant noise.  To generate the extracts, we paired each abstract
sentence with three sentences from the corresponding document,
selected using the techniques described by Marcu
\shortcite{marcu-sigir99}.  In an informal evaluation, $20$ such pairs
were randomly extracted and evaluated by a human.  Each pair was
ranked as $0$ (document sentences contain little-to-none of the
information in the abstract sentence), $1$ (document sentences contain
some of the information in the abstract sentence) or $2$ (document
sentences contain all of the information).  Of the twenty random
examples, none were labeled as $0$; five were labeled as $1$; and $15$
were labeled as $2$, giving a mean rating of $1.75$.  We refer to the
resulting corpus as the \extabs\ corpus, statistics for which
are shown in Table~\ref{fig:corpus-statistics2}.  Finally, for fair
comparison, we also run the Cut and Paste model only on the
extracts\footnote{\setlength{\baselineskip}{1\baselineskip}Interestingly, the Cut and Paste method actually
achieves \emph{higher} performance scores when run on only the
extracts rather than the full documents.}.

\begin{table}[t]
\tcaption{\setlength{\baselineskip}{1\baselineskip}Ziff-Davis extract corpus statistics}
\label{fig:corpus-statistics2}
\begin{tabular}{|l|r|r|r|}
\hline
&           {\bf Abstracts}  &    {\bf Extracts}    &    {\bf Documents}  \\
\hline
Documents           &\multicolumn{2}{c|}{$2033$} & $2033$ \\
Sentences           &   $13k$ &           $41k$ & $82k$ \\
Words               &  $261k$ &        $1m$ & $2.5m$ \\
\hline
Unique words               &   $14k$ &           $26k$ & $42k$ \\
                    &\multicolumn{2}{c|}{$29k$}  &  \\
\hline
Sentences/Doc       &     $6.28$ &            $21.51$ & $40.83$ \\
Words/Doc           &   $128.52$ &           $510.99$ & $1229.71$ \\
\hline
Words/Sent          &    $20.47$ &            $23.77$ & $28.36$ \\
\hline
\end{tabular}
\end{table}

\subsubsection{Machine translation models.}

We compare against several competing systems, the first of which is
based on the original IBM Model 4 for machine translation
\cite{brownetal93} and the HMM machine translation alignment model
\cite{vogel96} as implemented in the GIZA++ package \cite{och03}.  We
modified the code slightly to allow for longer inputs and higher
fertilities, but otherwise made no changes.  In all of these setups, 5
iterations of Model 1 were run, followed by 5 iterations of the HMM
model.  For Model 4, 5 iterations of Model 4 were subsequently run.

In our model, the distinction between the ``summary'' and the
``document'' is clear, but when using a model from machine
translation, it is unclear which of the summary and the document
should be considered the \emph{source} language and which should be
considered the \emph{target} language.  By making the summary the
source language, we are effectively requiring that the
\emph{fertility} of each summary word be very high, or that many words
are null generated (since we must generate all of the document).  By
making the document the source language, we are forcing the model to
make most document words have zero fertility.  We have performed
experiments in both directions, but the latter (document as source)
performs better in general.

In order to ``seed'' the machine translation model so that it knows
that word identity is a good solution, we appended our corpus with
``sentence pairs'' consisting of one ``source'' word and one
``target'' word, which were identical.  This is common practice in the
machine translation community when one wishes to cheaply encode
knowledge from a dictionary into the alignment model.

\subsubsection{Cut and Paste model.}

We also tested alignments using the Cut and Paste summary
decomposition method \cite{jing:cl}, based on a non-trainable HMM.
Briefly, the Cut and Paste HMM searches for long contiguous blocks of
words in the document and abstract that are identical (up to stem).
The longest such sequences are aligned.  By fixing a length cutoff of
$n$ and ignoring sequences of length less than $n$, one can
arbitrarily increase the precision of this method.  We found that
$n=2$ yields the best balance between precision and recall (and the
highest F-measure).  On this task, this model drastically outperforms
the machine translation models.

\subsubsection{The semi-Markov model.}

While the semi-HMM is based on a dynamic programming algorithm, the
effective search space in this model is enormous, even for moderately
sized \docabs\ pairs.  The semi-HMM system was then trained on this
\extabs\ corpus.  We also restrict the state-space with a beam, sized
at $50\%$ of the unrestricted state-space.  With this configuration,
we run ten iterations of the forward-backward algorithm.  The entire
computation time takes approximately 8 days on a 128 node cluster
computer.

We compare three settings of the semi-HMM.  The first, ``semi-HMM-relative''
uses the relative movement jump table; the second, ``semi-HMM-Gaussian''
uses the Gaussian parameterized jump table; the third,
``semi-HMM-syntax'' uses the syntax-based jump model.

\subsection{Evaluation results}

\begin{table}[tb]
\tcaption{\setlength{\baselineskip}{1\baselineskip}Results on the Ziff-Davis corpus}
\label{tab:results}
\begin{tabular}{|l|c|c|c||c|c|c|}
\hline
& \multicolumn{3}{c||}{\bf All Words} & \multicolumn{3}{c|}{\bf Non-Stop Words}\\
{\bf System} & 
{\bf SoftP} & {\bf Recall} & {\bf SoftF} &
{\bf SoftP} & {\bf Recall} & {\bf SoftF} \\
\hline
Human$_1$        &     0.727 &     0.746 & \bf 0.736 & \bf 0.751 & \bf 0.801 & \bf 0.775 \\
Human$_2$        &     0.680 &     0.695 &     0.687 &     0.730 &     0.722 &     0.726 \\
\hline
HMM (Sum=Src)    &     0.120 &     0.260 &     0.164 &     0.139 &     0.282 &     0.186 \\
Model 4 (Sum=Src)&     0.117 &     0.260 &     0.161 &     0.135 & \bf 0.283 &     0.183 \\
HMM (Doc=Src)    &     0.295 &     0.250 & \bf 0.271 & \bf 0.336 &     0.267 & \bf 0.298 \\
Model 4 (Doc=Src)&     0.280 &     0.247 &     0.262 &     0.327 &     0.268 &     0.295 \\
\hline
Cut \& Paste     &     0.349 &     0.379 & \bf 0.363 & \bf 0.431 & \bf 0.385 & \bf 0.407 \\
\hline 
semi-HMM-relative   &     0.456 &     0.686 &     0.548 &     0.512 &     0.706 &     0.593 \\
semi-HMM-Gaussian   &     0.328 &     0.573 &     0.417 &     0.401 &     0.588 &     0.477 \\
semi-HMM-syntax     &     0.504 &     0.701 & \bf 0.586 & \bf 0.522 & \bf 0.712 & \bf 0.606 \\
\hline
\end{tabular}
\end{table}

The results, in terms of precision, recall and f-score are shown in
Table~\ref{tab:results}.  The first three columns are when these three
statistics are computed over all words.  The next three columns are
when these statistics are only computed over words that do not appear
in our ``ignore list'' of 58 stop words.  Under the methodology for
combining the two human annotations by taking the union, either of the
human scores would achieve a precision and recall of $1.0$.  To give a
sense of how well humans actually perform on this task, we compare
each human against the other.

As we can see from Table~\ref{tab:results}, none of the machine
translation models is well suited to this task, achieving, at best, an
F-score of $0.298$.  The ``flipped'' models, in which the document
sentences are the ``source'' language and the abstract sentences are
the ``target'' language perform significantly better (comparatively).
Since the MT models are not symmetric, going the bad way requires that
many document words have zero fertility, which is difficult for these
models to cope with.

The Cut \& Paste method performs significantly better, which is to be
expected, since it is designed specifically for summarization.  As one
would expect, this method achieves higher precision than recall,
though not by very much.  The fact that the Cut \& Paste model
performs so well, compared to the MT models, which are able to learn
non-identity correspondences, suggests that any successful model
should be able to take advantage of both, as ours does.

Our methods significantly outperform both the IBM models and the Cut
\& Paste method, achieving a precision of $0.522$ and a recall of
$0.712$, yielding an overall F-score of $0.606$ when stop words are
not considered.  This is still below the human-against-human f-score
of $0.775$ (especially considering that the true human-against-human
scores are $1.0$), but significantly better than any of the other
models.

Among the three settings of our jump table, the syntax-based model
performs best, followed by the relative jump model, with the Gaussian
model coming in worst (though still better than any other approach).
Inspecting Figure~\ref{fig:movement-delta}, the fact that the Gaussian
model does not perform well is not surprising: the data shown there is
very non-Gaussian.  A double-exponential model might be a better fit,
but it is unlikely that such a model will outperform the syntax based
model, so we did not perform this experiment.

\subsection{Error analysis}


\begin{figure}[!bt]
{\bf Example 1:}\\
\mbox{\psfig{figure=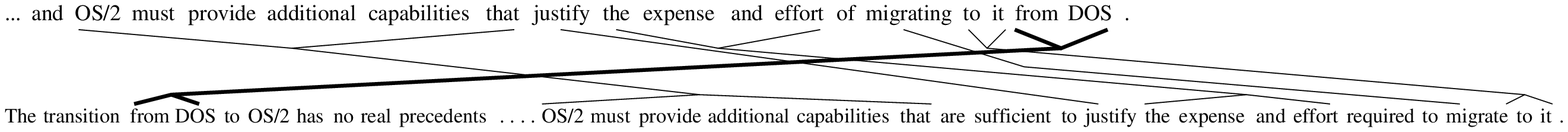,width=\textwidth,height=1.5cm}}

\mbox{~~~}

{\bf Example 2:}\\
\mbox{\psfig{figure=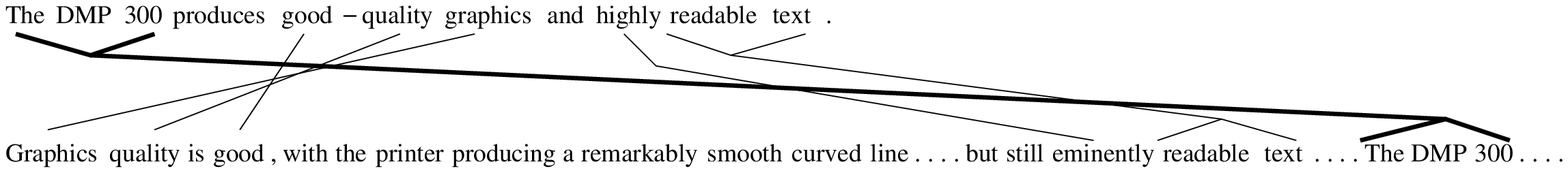,width=\textwidth,height=1.5cm}}

\caption{\setlength{\baselineskip}{1\baselineskip}Erroneous alignments are in bold.  (Top) Example of an error
  made by our model (from file ZF207-585-936).  ``From DOS'' should be
  null generated, but the model has erroneously aligned it to an
  identical phrase that appeared 11 sentences earlier in the document.
  (Bottom) Error (from ZF207-772-628); ``The DMP 300'' should be
  aligned to ``the printer'' but is instead aligned to a far-away
  occurrence of ``The DMP 300.''}
\label{fig:err-null}
\end{figure}

The first mistake frequently made by our model is to not align summary
words to null.  In effect, this means that our model of null-generated
summary words is lacking.  An example of this error is shown in
Example 1 in Figure~\ref{fig:err-null}.  In this example, the model
has erroneously aligned ``from DOS'' in the abstract to ``from DOS''
in the document (the error is shown in bold).  This alignment is wrong
because the context of ``from DOS'' in the document is completely
different from the context it appears in in the summary.  However, the
identity rewrite model has overwhelmed the locality model and forced
this incorrect alignment.  To measure the frequency of such errors, we
have post-processed our system's alignments so that whenever a human
alignment contains a null-generated summary word, our model also
predicts that this word is null-generated.  Doing so will not change
our system's recall, but it can improve the precision.  Indeed, in the
case of the relative jump model, the precision jumps from $0.456$ to
$0.523$ (f-score increases from $0.548$ to $0.594$) in the case of all
words and from $0.512$ to $0.559$ (f-score increases from $0.593$ to
$0.624$).  This corresponds to a relative improvement of roughly $8\%$
f-score.  Increases in score for the syntax-based model are roughly
the same.

\begin{table}[tb]
\tcaption{\setlength{\baselineskip}{1\baselineskip}10 example phrase alignments from the hand-annotated corpus;
  last column indicates whether the semi-HMM correctly aligned this phrase.}
\label{err2}
\begin{tabular}{|l|l|l|c|}
\hline
{\bf Summary Phrase} & {\bf True Phrase} & {\bf Aligned Phrase} & {\bf Class}\\
\hline
to port&                      can port               & to port  & incorrect \\
OS - 2&                       the OS / 2             & OS / 2   & partial \\
will use&                     will be using          & will using & partial \\
word processing programs&     word processors        & word processing & incorrect \\
consists of&                  also includes          & \emph{null} of & partial \\
will test&                    will also have to test & will test & partial \\
the potential buyer&          many users             & the buyers & incorrect \\
The new software&             Crosstalk for Windows  & new software & incorrect \\
are generally powered by&     run on                 & \emph{null} & incorrect \\
Oracle Corp.&                 the software publisher & Oracle Corp. & incorrect \\
\hline
\end{tabular}
\end{table}

The second mistake our model frequently makes is to trust the identity
rewrite model too strongly.  This problem has to do either with
synonyms that do not appear frequently enough for the system to learn
reliable rewrite probabilities, or with coreference issues, in which
the system chooses to align, for instance, ``Microsoft'' to
``Microsoft,'' rather than ``Microsoft'' to ``the company,'' as might
be correct in context.  As suggested by this example, this problem is
typically manifested in the context of coreferential noun phrases.  It
is difficult to perform a similar analysis of this problem as for the
aforementioned problem (to achieve an upper bound on performance), but
we can provide some evidence.  As mentioned before, in the human
alignments, roughly $51\%$ of all aligned phrases are lexically
identical.  In the alignments produced by our model (on the same
documents), this number is $69\%$.  In the case of stem identity, the
hand align data suggests that stem identity should hold in $67\%$ of
the cases; in our alignments, this number was $81\%$.  An example of
this sort of error is shown in Example 2 in Figure~\ref{fig:err-null}.
Here, the model has aligned ``The DMP 300'' in the abstract to ``The
DMP 300'' in the document, while it should have been aligned to ``the
printer'' due to locality constraints (note that the model also misses
the (produces $\leftrightarrow$ producing) alignment, likely as a
side-effect of it making the error depicted in bold).

In Table~\ref{err2}, we have shown examples of common errors made by
our system (these were randomly selected from a much longer list of
errors).  These examples are shown out of their contexts, but in most
cases, the error is clear even so.  In the first column, we show the
summary phrase in question.  In the second column, we show the
document phrase do which it \emph{should} be aligned, and in the third
column, we show the document phrase that our model aligned it to (or
``\emph{null}'').  In the right column, we classify the model's
alignment as ``incorrect'' or ``partially correct.''

The errors shown in Table~\ref{err2} show several weaknesses of the
model.  For instance, in the first example, it aligns ``to port'' with
``to port,'' which seems correct without context, but the chosen
occurrence of ``to port'' in the document is in the discussion of a
completely different porting process than that referred to in the
summary (and is several sentences away).  The seventh and tenth
examples (``The new software'' and ``Oracle Corp.'' respectively) show
instances of the coreference error that occurs commonly.

\section{Conclusion and Discussion} \label{sec:conclusion}

Currently, summarization systems are limited to either using
hand-annotated data, or using weak alignment models at the granularity
of sentences, which serve as suitable training data only for sentence
extraction systems.  To train more advanced extraction systems, such
as those used in document compression models or in next-generation
abstraction models, we need to better understand the lexical
correspondences between documents and their human written abstracts.
Our work is motivated by the desire to leverage the vast number of
\docabs\ pairs that are freely available on the internet and in other
collections, and to create word- and phrase-aligned \docabs\ corpora
automatically.

This paper presents a statistical model for learning such alignments
in a completely unsupervised manner.  The model is based on an
extension of a hidden Markov model, in which multiple emissions are
made in the course of one transition.  We have described efficient
algorithms in this framework, all based on dynamic programming.  Using
this framework, we have experimented with \emph{complex} models of
movement and lexical correspondences.  Unlike the approaches used in
machine translation, where only very simple models are used, we have
shown how to efficiently and effectively leverage such disparate
knowledge sources and WordNet, syntax trees and identity models.

We have empirically demonstrated that our model is able to learn the
complex structure of \docabs\ pairs.  Our system outperforms
competing approaches, including the standard machine translation
alignment models \cite{brownetal93,vogel96} and the state of the art
Cut \& Paste summary alignment technique \cite{jing:cl}.  

We have analyzed two sources of error in our model, including issues
of null-generated summary words and lexical identity.  Within the
model itself, we have already suggested two major sources of error in
our alignment procedure.  Clearly more work needs to be done to fix
these problems.  One approach that we believe will be particularly
fruitful would be to add a fifth model to the linearly interpolated
rewrite model based on lists of synonyms automatically extracted from
large corpora.  Additionally, investigating the possibility of
including some sort of weak coreference knowledge into the model might
serve to help with the second class of errors made by the model.

One obvious aspect of our method that may reduce its general
usefulness is the computation time.  In fact, we found that despite
the efficient dynamic programming algorithms available for this model,
the state space and output alphabet are simply so large and complex
that we were forced to first map documents down to extracts before we
could process them (and even so, computation took roughly 1000
processor hours).  Though we have not done so in this work, we do
believe that there is room for improvement computationally, as well.
One obvious first approach would be to run a simpler model for the
first iteration (for example, Model 1 from machine translation
\cite{brownetal93}, which tends to be very ``recall oriented'') and
use this to see subsequent iterations of the more complex model.  By
doing so, one could \emph{recreate} the extracts at each iteration
using the previous iteration's parameters to make better and
\emph{shorter} extracts.  Similarly, one might only allow summary
words to align to words found in their corresponding extract
sentences, which would serve to significantly speed up training and,
combined with the parameterized extracts, might not hurt performance.
A final option, but one that we do not advocate, would be to give up
on phrases and train the model in a word-to-word fashion.  This could
be coupled with heuristic phrasal creation as is done in machine
translation \cite{och00alignment}, but by doing so, one completely
loses the probabilistic interpretation that makes this model so
pleasing.

Aside from computational considerations, the most obvious future
effort along the vein of this model is to incorporate it into a full
document summarization system.  Since this can be done in many ways,
including training extraction systems, compression systems, headline
generation systems and even extraction systems, we leave this to
future work so that we could focus specifically on the alignment task
in this paper.  Nevertheless, the true usefulness of this model will
be borne out by its application to true summarization tasks.

\starttwocolumn

\begin{acknowledgments}

\setlength{\baselineskip}{1\baselineskip}      

We wish to thank David Blei for helpful theoretical discussions
related to this project and Franz Josef Och for sharing his technical
expertise on issues that made the computations discussed in this paper
possible.  We sincerely thank the anonymous reviewers of an original
conference version of this article as well reviewers of this longer
version, all of whom gave very useful suggestions.  Some of the
computations described in this work were made possible by the High
Performance Computing Center at the University of Southern California.
This work was partially supported by DARPA-ITO grant N66001-00-1-9814,
NSF grant IIS-0097846, NSF grant IIS-0326276, and a USC Dean
Fellowship to Hal Daum\'e III.

\end{acknowledgments}

\small
\bibliographystyle{fullname}
\setlength{\baselineskip}{1\baselineskip}      
\bibliography{bibfile}

\begin{thebibliography}{}

\bibitem[\protect\citename{Aydin, Altunbasak, and
  Borodovsky}2004]{aydin04protein}
Aydin, Zafer, Yucel Altunbasak, and Mark Borodovsky.
\newblock 2004.
\newblock Protein secondary structure prediction with semi-{Markov} {HMM}s.
\newblock In {\em Proceedings of the IEEE International Conference on
  Acoustics, Speech and Signal Processing (ICASSP)}, May.

\bibitem[\protect\citename{Banko, Mittal, and Witbrock}2000]{bankoetal00}
Banko, Michele, Vibhu Mittal, and Michael Witbrock.
\newblock 2000.
\newblock Headline generation based on statistical translation.
\newblock In {\em Proceedings of the Conference of the Association for
  Computational Linguistics (ACL)}, pages 318--325, Hong Kong, October 1--8.

\bibitem[\protect\citename{Barzilay and Elhadad}2003]{barzilay03alignment}
Barzilay, Regina and Noemie Elhadad.
\newblock 2003.
\newblock Sentence alignment for monolingual comparable corpora.
\newblock In {\em Proceedings of the Conference on Empirical Methods in Natural
  Language Processing (EMNLP)}, pages 25--32.

\bibitem[\protect\citename{Baum and Eagon}1967]{baum67hmms}
Baum, Leonard~E. and J.E. Eagon.
\newblock 1967.
\newblock An inequality with applications to statistical estimation for
  probabilistic functions of {Markov} processes and to a model of ecology.
\newblock {\em Bulletins of the American Mathematical Society}, 73:360--363.

\bibitem[\protect\citename{Baum and Petrie}1966]{baum66hmms}
Baum, Leonard~E. and Ted Petrie.
\newblock 1966.
\newblock Statistical inference for probabilistic functions of finite state
  {Markov} chains.
\newblock {\em Annals of Mathematical Statistics}, 37:1554--1563.

\bibitem[\protect\citename{Berger and Mittal}2000]{berger-mittal00}
Berger, Adam and Vibhu Mittal.
\newblock 2000.
\newblock Query-relevant summarization using {FAQ}s.
\newblock In {\em Proceedings of the Conference of the Association for
  Computational Linguistics (ACL)}, pages 294--301, Hong Kong, October 1--8.

\bibitem[\protect\citename{Boyles}1983]{boyles83em}
Boyles, Russell~A.
\newblock 1983.
\newblock On the convergence of the {EM} algorithm.
\newblock {\em Journal of the Royal Statistical Society}, B(44):47--50.

\bibitem[\protect\citename{Brown \bgroup et al.\egroup }1993]{brownetal93}
Brown, {Peter F.}, {Stephen A.} {Della Pietra}, {Vincent J.} {Della Pietra},
  and {Robert L.} Mercer.
\newblock 1993.
\newblock The mathematics of statistical machine translation: Parameter
  estimation.
\newblock {\em Computational Linguistics}, 19(2):263--311.

\bibitem[\protect\citename{Carletta}1995]{carletta96kappa}
Carletta, Jean.
\newblock 1995.
\newblock Assessing agreement on classification tasks: the kappa statistic.
\newblock {\em Computational Linguistics}, 22(2):249--254.

\bibitem[\protect\citename{Carroll \bgroup et al.\egroup }1998]{carrolletal98}
Carroll, John, Guido Minnen, Yvonne Canning, Siobhan Devlin, and John Tait.
\newblock 1998.
\newblock Practical simplification of {E}nglish newspaper text to assist
  aphasic readers.
\newblock In {\em Proceedings of the AAAI-98 Workshop on Integrating Artificial
  Intelligence and Assistive Technology}.

\bibitem[\protect\citename{Chandrasekar, Doran, and
  Bangalore}1996]{chandrasekaretal96}
Chandrasekar, Raman, Christy Doran, and Srinivas Bangalore.
\newblock 1996.
\newblock Motivations and methods for text simplification.
\newblock In {\em Proceedings of the International Conference on Computational
  Linguistics (COLING)}, pages 1041--1044, Copenhagen, Denmark.

\bibitem[\protect\citename{Charniak}1997]{charniak97}
Charniak, Eugene.
\newblock 1997.
\newblock Statistical parsing with a context-free grammar and word statistics.
\newblock In {\em Proceedings of the Fourteenth National Conference on
  Artificial Intelligence (AAAI'97)}, pages 598--603, Providence, Rhode Island,
  July 27--31.

\bibitem[\protect\citename{{Daum\'e III} and Marcu}2002]{daume02noisychannel}
{Daum\'e III}, Hal and Daniel Marcu.
\newblock 2002.
\newblock A noisy-channel model for document compression.
\newblock In {\em Proceedings of the Conference of the Association for
  Computational Linguistics (ACL)}, pages 449--456.

\bibitem[\protect\citename{{Daum\'e III} and Marcu}2004]{daume04treeposition}
{Daum\'e III}, Hal and Daniel Marcu.
\newblock 2004.
\newblock A tree-position kernel for document compression.
\newblock In {\em Proceedings of the Fourth Document Understanding Conference
  (DUC 2004)}, Boston, MA, May 6 -- 7.

\bibitem[\protect\citename{Dempster, Laird, and Rubin}1977]{dempster77em}
Dempster, A.P., N.M. Laird, and D.B. Rubin.
\newblock 1977.
\newblock Maximum likelihood from incomplete data via the {EM} algorithm.
\newblock {\em Journal of the Royal Statistical Society}, B39.

\bibitem[\protect\citename{Edmundson}1969]{edmundson69}
Edmundson, {H.P.}
\newblock 1969.
\newblock New methods in automatic abstracting.
\newblock {\em Journal of the Association for Computing Machinery},
  16(2):264--285.
\newblock Reprinted in \emph{Advances in Automatic Text Summarization}, I. Mani
  and M.T. Maybury, (eds.).

\bibitem[\protect\citename{Eisner}2003]{eisner03treemappings}
Eisner, Jason.
\newblock 2003.
\newblock Learning non-isomorphic tree mappings for machine translation.
\newblock In {\em Proceedings of the Conference of the Association for
  Computational Linguistics (ACL)}.

\bibitem[\protect\citename{Ferguson}1980]{ferguson80variable}
Ferguson, Jack~D.
\newblock 1980.
\newblock Variable duration models for speech.
\newblock In {\em Proceedings of the Symposium on the Application of Hidden
  {Markov} Models to Text and Speech}, pages 143--179, October.

\bibitem[\protect\citename{Gales and Young}1993]{gales93segmental}
Gales, Mark~J.F. and Steve~J. Young.
\newblock 1993.
\newblock The theory of segmental hidden {Markov} models.
\newblock Technical report, Cambridge University Engineering Department.

\bibitem[\protect\citename{Gauvain and Lee}1994]{gauvain94maximum}
Gauvain, Jean-Luc and Chin-Hui Lee.
\newblock 1994.
\newblock Maximum a-posteriori estimation for multivariate gaussian mixture
  observations of {Markov} chains.
\newblock {\em IEEE Transactions Speech and Audio Processing}, 2:291--298.

\bibitem[\protect\citename{Ge and Smyth}2000]{ge00segmental}
Ge, Xianping and Padhraic Smyth.
\newblock 2000.
\newblock Segmental semi-{Markov} models for change-point detection with
  applications to semiconductor manufacturing.
\newblock Technical report, University of California at Irvine, March.

\bibitem[\protect\citename{Gildea}2003]{gildea03alignment}
Gildea, Daniel.
\newblock 2003.
\newblock Loosely tree-based alignment for machine translation.
\newblock In {\em Proceedings of the Conference of the Association for
  Computational Linguistics (ACL)}, pages 80--87.

\bibitem[\protect\citename{Grefenstette}1998]{grefenstette98}
Grefenstette, Gregory.
\newblock 1998.
\newblock Producing intelligent telegraphic text reduction to provide an audio
  scanning service for the blind.
\newblock In {\em Working Notes of the AAAI Spring Symposium on Intelligent
  Text Summarization}, pages 111--118, Stanford University, CA, March 23-25.

\bibitem[\protect\citename{Jensen}1906]{jensen06inequality}
Jensen, J.L.W.V.
\newblock 1906.
\newblock Sur les fonctions convexes et les in\'egalit\'es entre les valeurs
  moyennes.
\newblock {\em Acta Mathematica}, 30:175--193.

\bibitem[\protect\citename{Jing}2000]{jing00}
Jing, Hongyan.
\newblock 2000.
\newblock Sentence reduction for automatic text summarization.
\newblock In {\em Proceedings of the Conference of the North American Chapter
  of the Association for Computational Linguistics (NAACL)}, pages 310--315,
  Seattle, WA.

\bibitem[\protect\citename{Jing}2002]{jing:cl}
Jing, Hongyan.
\newblock 2002.
\newblock Using hidden markov modeling to decompose human-written summaries.
\newblock {\em Computational Linguistics}, 28(4):527 -- 544, December.

\bibitem[\protect\citename{Knight and Marcu}2002]{knight-marcu02}
Knight, Kevin and Daniel Marcu.
\newblock 2002.
\newblock Summarization beyond sentence extraction: A probabilistic approach to
  sentence compression.
\newblock {\em Artificial Intelligence}, 139(1).

\bibitem[\protect\citename{Kupiec, Pedersen, and Chen}1995]{kupiec95trainable}
Kupiec, Julian, Jan~O. Pedersen, and Francine Chen.
\newblock 1995.
\newblock A trainable document summarizer.
\newblock In {\em Research and Development in Information Retrieval}, pages
  68--73.

\bibitem[\protect\citename{Luhn}1956]{luhn56}
Luhn, H.P.
\newblock 1956.
\newblock The automatic creation of literature abstracts.
\newblock In Inderjeet Mani and Mark Maybury, editors, {\em Advances in
  Automatic Text Summarization}. The MIT Press, Cambridge, MA, pages 58--63.

\bibitem[\protect\citename{Mahesh}1997]{mahesh97}
Mahesh, Kavi.
\newblock 1997.
\newblock Hypertext summary extraction for fast document browsing.
\newblock In {\em Proceedings of the AAAI Spring Symposium on Natural Language
  Processing for the World Wide Web}, pages 95--103.

\bibitem[\protect\citename{Mani}2001]{mani-book01}
Mani, Inderjeet.
\newblock 2001.
\newblock {\em Automatic Summarization}, volume~3 of {\em Natural Language
  Processing}.
\newblock John Benjamins Publishing Company, Amsterdam/Philadelphia.

\bibitem[\protect\citename{Mani and Maybury}1999]{mani-maybury-book99}
Mani, Inderjeet and Mark Maybury, editors.
\newblock 1999.
\newblock {\em Advances in Automatic Text Summarization}.
\newblock The MIT Press, Cambridge, MA.

\bibitem[\protect\citename{Marcu}1999]{marcu-sigir99}
Marcu, Daniel.
\newblock 1999.
\newblock The automatic construction of large-scale corpora for summarization
  research.
\newblock In {\em Proceedings of the 22nd Conference on Research and
  Development in Information Retrieval (SIGIR--99)}, pages 137--144, Berkeley,
  CA, August 15--19.

\bibitem[\protect\citename{Marcu}2000]{marcu-book00}
Marcu, Daniel.
\newblock 2000.
\newblock {\em The Theory and Practice of Discourse Parsing and Summarization}.
\newblock The MIT Press, Cambridge, Massachusetts.

\bibitem[\protect\citename{Mitchell, Jamieson, and
  Harper}1995]{mitchell95duration}
Mitchell, Carl~D., Leah~H. Jamieson, and Mary~P. Harper.
\newblock 1995.
\newblock On the complexity of explicit duration {HMM}s.
\newblock {\em IEEE Transactions on Speech and Audio Processing}, 3(3), May.

\bibitem[\protect\citename{Och and Ney}2000]{och00alignment}
Och, Franz~Josef and Hermann Ney.
\newblock 2000.
\newblock Improved statistical alignment models.
\newblock In {\em Proceedings of the Conference of the Association for
  Computational Linguistics (ACL)}, pages 440--447, October.

\bibitem[\protect\citename{Och and Ney}2003]{och03}
Och, Franz~Josef and Hermann Ney.
\newblock 2003.
\newblock A systematic comparison of various statistical alignment models.
\newblock {\em Computational Linguistics}, 29(1):19--51.

\bibitem[\protect\citename{Ostendorf, Digalakis, and
  Kimball}1996]{ostendorf96segment}
Ostendorf, Mari, Vassilis Digalakis, and Owen Kimball.
\newblock 1996.
\newblock From {HMM}s to segment models: A unified view of stochastic modeling
  for speech recognition.
\newblock {\em IEEE Transactions on Speech and Audio Processing},
  4(5):360--378, September.

\bibitem[\protect\citename{Quirk, Brockett, and Dolan}2004]{quirk04alignment}
Quirk, Chris, Chris Brockett, and William Dolan.
\newblock 2004.
\newblock Monolingual machine translation for paraphrase generation.
\newblock In {\em Proceedings of the Conference on Empirical Methods in Natural
  Language Processing (EMNLP)}, pages 142--149, Barcelona, Spain.

\bibitem[\protect\citename{Schwartz, Zajic, and Dorr}2002]{schwartz-headline}
Schwartz, Richard, David Zajic, and Bonnie Dorr.
\newblock 2002.
\newblock Automatic headline generation for newspaper stories.
\newblock In {\em Proceedings of the Document Understanding Conference (DUC)},
  pages 78--85.

\bibitem[\protect\citename{Smyth, Heckerman, and Jordan}1997]{smyth97pins}
Smyth, Padhraic, David Heckerman, and Michael~I. Jordan.
\newblock 1997.
\newblock Probabilistic independence networks for hidden {Markov} probability
  models.
\newblock {\em Neural Computation}, 9(2):227--269.

\bibitem[\protect\citename{Teufel and Moens}1997]{teufel97sentence}
Teufel, Simone and Mark Moens.
\newblock 1997.
\newblock Sentence extraction as a classification task.
\newblock In {\em In ACL/EACL-97 Workshop on Intelligent and Scalable Text
  Summarization}, pages 58--65.

\bibitem[\protect\citename{Vogel, Ney, and Tillmann}1996]{vogel96}
Vogel, Stephan, Hermann Ney, and Christoph Tillmann.
\newblock 1996.
\newblock {HMM}-based word alignment in statistical translation.
\newblock In {\em Proceedings of the International Conference on Computational
  Linguistics (COLING)}, pages 836--841.

\bibitem[\protect\citename{Wu}1983]{wu83em}
Wu, Jeff~C.F.
\newblock 1983.
\newblock On the convergence properties of the {EM} algorithm.
\newblock {\em The Annals of Statistics}, 11:95--103.

\bibitem[\protect\citename{Yamada and Knight}2001]{yamada01}
Yamada, Kenji and Kevin Knight.
\newblock 2001.
\newblock A syntax-based statistical translation model.
\newblock In {\em Proceedings of the Conference of the Association for
  Computational Linguistics (ACL)}, pages 523--530.

\bibitem[\protect\citename{Zajic, Dorr, and Schwartz}2004]{zajic04topiary}
Zajic, David, Bonnie Dorr, and Richard Schwartz.
\newblock 2004.
\newblock {BBN/UMD} at {DUC}-2004: {T}opiary.
\newblock In {\em Proceedings of the Fourth Document Understanding Conference
  (DUC 2004)}, Boston, MA, May 6 -- 7.

\end{thebibliography}

\end{document}